\documentclass[letterpaper]{article} % DO NOT CHANGE THIS
\usepackage{arxiv_aaai2026}  % DO NOT CHANGE THIS
\usepackage{times}  % DO NOT CHANGE THIS
\usepackage{helvet}  % DO NOT CHANGE THIS
\usepackage{courier}  % DO NOT CHANGE THIS
\usepackage[hyphens]{url}  % DO NOT CHANGE THIS
\usepackage{graphicx} % DO NOT CHANGE THIS
\usepackage{natbib}  % DO NOT CHANGE THIS AND DO NOT ADD ANY OPTIONS TO IT
\usepackage{caption} % DO NOT CHANGE THIS AND DO NOT ADD ANY OPTIONS TO IT
\usepackage{algorithm}
\usepackage{algorithmic}
\nocopyright

\usepackage{newfloat}
\usepackage{listings}
\DeclareCaptionStyle{ruled}{labelfont=normalfont,labelsep=colon,strut=off} % DO NOT CHANGE THIS
\floatstyle{ruled}
\newfloat{listing}{tb}{lst}{}
\floatname{listing}{Listing}
\title{Do Large Language Models Truly Understand Cross-cultural Differences?}
\author{
  Shiwei Guo\textsuperscript{\rm 1},
  Sihang Jiang\textsuperscript{\rm 2},
  Qianxi He\textsuperscript{\rm 2},
  Yanghua Xiao\textsuperscript{\rm 2},\thanks{Corresponding author.}
  Jiaqing Liang\textsuperscript{\rm 2},
  Bi Yude\textsuperscript{\rm 1},
  Minggui He\textsuperscript{\rm 3},
  Shimin Tao\textsuperscript{\rm 3},
  Li Zhang\textsuperscript{\rm 3}
}

\affiliations{
  \textsuperscript{\rm 1}School of Foreign Languages and Literatures, Fudan University\\
  \textsuperscript{\rm 2}Shanghai Key Laboratory of Data Science, College of Computer Science and Artificial Intelligence, Fudan University\\
  \textsuperscript{\rm 3}Huawei, China\\
}

\begin{document}

\maketitle

\begin{abstract}
In recent years, large language models (LLMs) have demonstrated strong performance on multilingual tasks. Given its wide range of applications, cross-cultural understanding capability is a crucial competency. However, existing benchmarks for evaluating whether LLMs genuinely possess this capability suffer from three key limitations: a lack of contextual scenarios, insufficient cross-cultural concept mapping, and limited deep cultural reasoning capabilities. To address these gaps, we propose SAGE, a scenario-based benchmark built via cross-cultural core concept alignment and generative task design, to evaluate LLMs' cross-cultural understanding and reasoning. Grounded in cultural theory, we categorize cross-cultural capabilities into nine dimensions. Using this framework, we curated 210 core concepts and constructed 4530 test items across 15 specific real-world scenarios, organized under four broader categories of cross-cultural situations, following established item design principles. The SAGE dataset supports continuous expansion, and experiments confirm its transferability to other languages. It reveals model weaknesses across both dimensions and scenarios, exposing systematic limitations in cross-cultural reasoning. While progress has been made, LLMs are still some distance away from reaching a truly nuanced cross-cultural understanding. In compliance with the anonymity policy, we include data and code in the supplement materials. In future versions, we will make them publicly available online.
\end{abstract}

\section{Introduction}

With the continued advancement of LLMs, researchers have increasingly incorporated multilingual data~\cite{singh2024translating} to enhance the models’ abilities across diverse languages~\cite{lai2024xllms100}. This progress has significantly improved LLM performance in tasks such as translation~\cite{wang2025multilingualprompting}, cross‑border transactions~\cite{li2024culturellm}, and international communication~\cite{taori2023stanford}. Despite these improvements, researchers increasingly acknowledge that cross-cultural understanding is a critical competence that goes beyond multilingual capability~\cite{rystrom2025multilingual} and is essential for the effective deployment of LLMs in real-world contexts~\cite{havaldar2023multilingual}. As a result, \textbf{cross-cultural understanding} is increasingly regarded as a key ability and evaluation target~\cite{cao2025mceval} for LLMs, especially in culturally diverse and semantically nuanced scenarios~\cite{liu2025culturevlm}.

\begin{figure}[t]
\centering
\includegraphics[width=0.95\columnwidth]{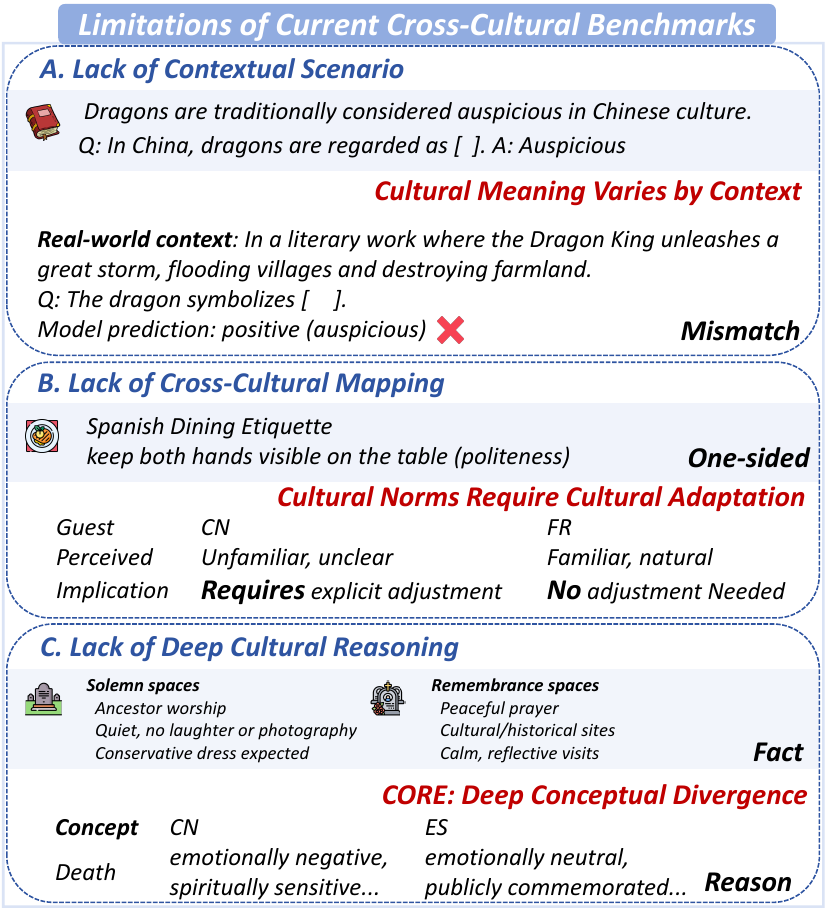}
\caption{Cross-cultural understanding is context-sensitive, necessitates cultural adaptation, and involves grasping the underlying values and cognitive frameworks behind cultural behaviors.}
\label{fig 1}
\end{figure}

Existing research primarily focus on cultural commonsense knowledge to address questions, as shown in Fig.~\ref{fig 1}. Including commonly observed practices related to food~\cite{palta2023fork}, behavior~\cite{yin2022geomlama}, and etiquette~\cite{aroca2023candle} within specific cultures. By incorporating this type of knowledge~\cite{nguyen2024mango}, LLMs are expected to demonstrate basic cultural recognition~\cite{li2023geolm} and adaptability~\cite{zhao2025xcsr} across diverse cultural settings~\cite{kargaran2024m4}. 

To advance cross-cultural understanding evaluation of LLMs, we identify three critical limitations in current approaches. First, \textbf{Lack of cross-cultural contextual scenarios}: While LLMs may store factual cultural knowledge~\cite{li2024culturegen}, they often struggle to apply this knowledge appropriately in real-world settings. As illustrated in Fig.~\ref{fig 1}A, cultural fact may yield different interpretations across situations. Without grounded scenarios, evaluations fail to capture models’ ability to reason within culturally situated contexts~\cite{deardorff2006identification}. 

Second, \textbf{Lack of cross-cultural mapping}: Existing work often collects culture-specific knowledge as standalone facts, such as dining etiquette rules for a single country, but does not map how people from other cultural backgrounds would interpret and respond to these norms. As shown in Fig.~\ref{fig 1}B, a model may accurately recite Spanish dining etiquette, yet fail to anticipate what a Chinese or French guest would find most salient or potentially problematic in the same setting. Cross-cultural understanding therefore requires aligning expectations, interpretations, and responses across cultures, rather than one-sided fact recall~\cite{tao2024cultural,nishida1999cognitive,spencer2008culturally}.

Third, \textbf{Lack of deep cultural reasoning}: Even when LLMs acquire multilingual cultural facts~\cite{alkhamissi2024investigating}, their outputs often remain superficial. In practice, surface cultural knowledge can be endlessly expanded, but fact accumulation alone does not tell a model \emph{why} a concept matters in a given culture. As illustrated in Fig.~\ref{fig 1}C, terms such as 'cemetery' may be known across languages~\cite{xu2025does}, yet models still fail to reason about the culturally grounded symbolic meanings and emotional norms they evoke~\cite{shen2024understanding, zhou2025culture}. What is missing is reasoning anchored in deeper cultural schemas and conceptual frameworks, so that models can not only state what is done, but also explain why it is done~\cite{shaules2007deep, hall1976beyond}.

To address these key limitations presents several challenges.
First, real-world cross-cultural contexts are inherently complex and variable~\cite{rauba2024context}, as cultural knowledge~\cite{triandis1993collectivism} can manifest differently depending on the scenario~\cite{sun2025casebench}.
Second, cross-cultural understanding requires dynamic adaptation across cultures~\cite{bu2025heritage}, as it involves interactions between cultural systems. Third, the core challenge lies in that understanding the causes of cultural differences requires more than the collection of surface-level knowledge~\cite{hall1976beyond}. It involves modeling deep cultural structures~\cite{wittgenstein1968philosophical}, such as values, cognitive patterns, and communicative norms, which are often implicit and difficult to observe.

\begin{figure*}[!ht]
\centering
\includegraphics[width=0.95\textwidth]{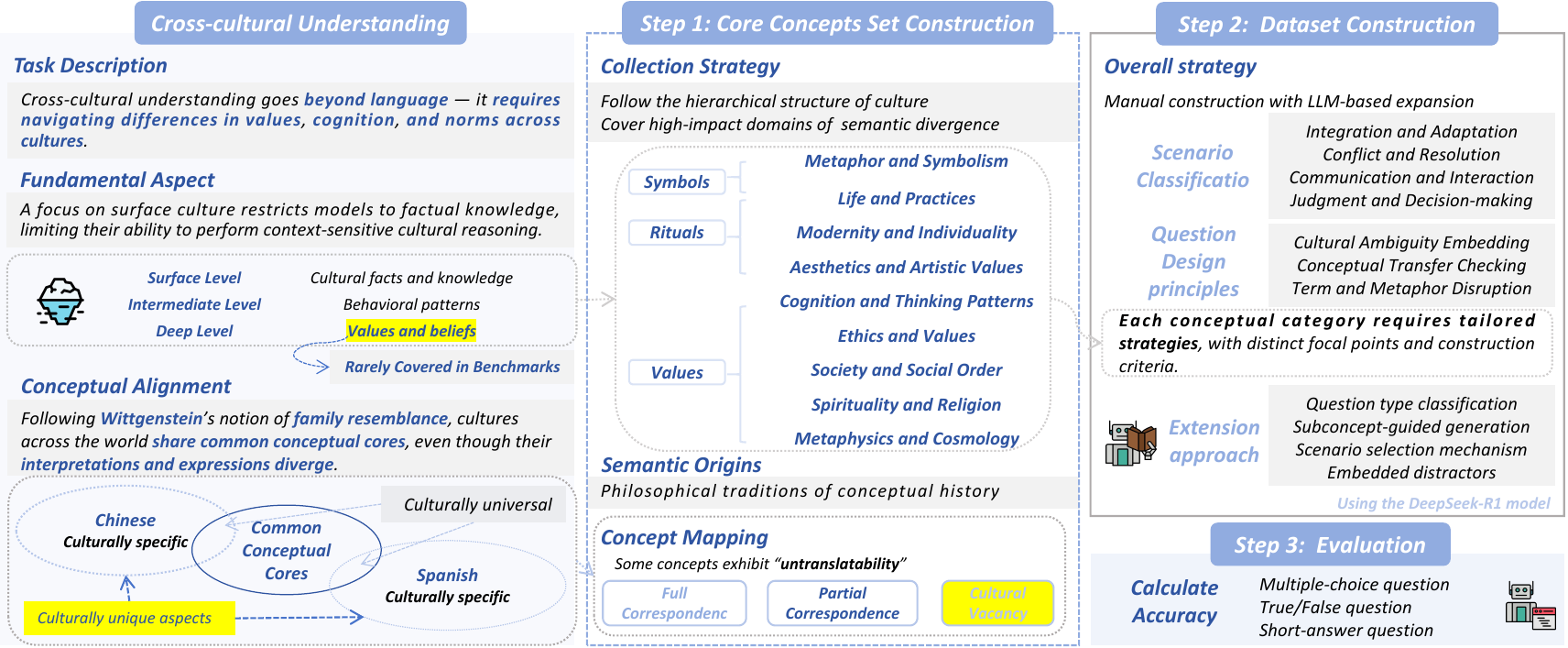} 
\caption{SAGE Benchmark Design Framework.}
\label{fig 2}
\end{figure*}

Therefore, we propose the \textbf{SAGE} benchmark, a \underline{\textbf{S}}cenario-based evaluation framework constructed through cross-cultural core concept \underline{\textbf{A}}lignment and \underline{\textbf{GE}}nerative task design, aiming to assess LLMs’ understanding and reasoning abilities. The framework of
our benchmark is shown in Fig.~\ref{fig 2}. Unlike evaluations built on externally chosen culture facts, SAGE is grounded in \textbf{Cross-cultural Core Concepts} (CCCs), a technical term from cultural studies and conceptual history. CCCs are defined and refined by scholars within each culture, so their meanings reflect local interpretive traditions rather than outsider imposition. Importantly, CCCs are dynamic rather than fixed: their senses shift with historical discourse and social practice, which makes them suitable for testing culturally situated interpretation instead of static recall. Our design also explicitly handles \textbf{Cultural Vacancy}, where a concept is salient in one culture but lacks a true counterpart in another; we preserve such asymmetries rather than forcing one-to-one mappings, enabling culturally authentic and scalable multilingual extension.

To operationalize CCCs, we first construct a CCC set guided by broad theoretical foundations (see Appendix~A) and derive a three-level layers~\cite{trompenaars1998riding} comprising nine major cross-cultural categories~\cite{unesco2013intercultural}, yielding 210 concepts per language. Building on this CCC set, we further construct the SAGE evaluation benchmark by classifying real-world cultural interactions~\cite{hammer2003measuring} into four contextual scenario types and fifteen cross-cultural communication scenarios, and integrating CCCs into context-driven tasks, resulting in 4,530 culturally grounded questions. To support future growth, we provide dataset expansion principles. We select Chinese and Spanish as initial target languages to reduce Anglophone dominance, leverage large speaker bases, and ensure linguistic diversity; these \textbf{pilot languages} also support subsequent high-quality extensions to \textbf{low-resource languages} via minimal human intervention and rule-based generation.

The contributions are summarized as follows:
\begin{itemize}
    \item \textbf{First methodology} using Cross-Cultural Core Concept Sets to evaluate LLMs' cross-cultural understanding capabilities.
    \item \textbf{SAGE benchmark} integrating cultural core concepts with real-world scenarios for authentic cross-cultural capability assessment.
    \item \textbf{Scenario-based task suite} intentionally avoiding stereotypes and monocultural assumptions, spanning 4 pragmatic interaction types across 15 culturally diverse situations.
    \item \textbf{Extensible framework} supporting continuous expansion and seamless transfer to additional languages through minimal manual effort.
\end{itemize}

\section{SAGE Benchmark}

\subsection{Cross-cultural Core Concept Set}

\paragraph{Collection Strategy}
We operationalize CCCs with a reproducible taxonomy that organizes cultural content into \textbf{three layers} (\textit{Symbolic}, \textit{Behavioral \& Ritual}, \textit{Value}) and \textbf{nine categories} (Fig.~\ref{fig 2}): (1) \textbf{Metaphor and Symbolism}, (2) \textbf{Life and Practices}, (3) \textbf{Modernity and Individuality}, (4) \textbf{Aesthetics and Artistic Values}, (5) \textbf{Cognition and Thinking Patterns}, (6) \textbf{Ethics and Values}, (7) \textbf{Society and Social Order}, (8) \textbf{Spirituality and Religion}, and (9) \textbf{Metaphysics and Cosmology}. This taxonomy is designed to cover recurrent domains where cross-cultural misunderstandings and interpretation gaps frequently arise in intercultural communication, providing a structured space for concept collection and evaluation. A detailed explanation of the taxonomy and its theoretical grounding is provided in \textbf{Appendix~A}. 

\paragraph{Selection Protocol}
We adopt a two-stage strategy: (full protocol and released artifacts are detailed in Appendix~B)
\begin{itemize}
  \item \textbf{Stage I (shared CCC pool).} We first collect a candidate pool of cross-culturally recurrent concepts under the CCC taxonomy (Fig.~\ref{fig 2}). Candidates are compiled by experts and stored as concept cards with (i) category label(s), (ii) a short definition, (iii) usage notes, and (iv) at least one authoritative scholarly source.
  \item \textbf{Stage II (culture-specific CCC pool).} We then add culture-specific concepts~\cite{wierzbicka1997understanding} that capture values, institutions, or experiences salient in a particular cultural context. In this stage, a large language model is used only as a constrained candidate generator/retriever: given the taxonomy and a fixed prompt, it proposes candidates with evidence snippets and mapping notes, while experts make the final inclusion decision.
\end{itemize}

Candidates are retained only if they satisfy the following inclusion criteria (see Appendix~B):
\begin{itemize}
  \item \textbf{Cultural salience}: attested in authoritative cultural/historical scholarship rather than transient topics;
  \item \textbf{Conceptual specificity}: a stable semantic unit suitable for operationalization;
  \item \textbf{Cross-cultural evaluability}: assessable via scenario-based judgments or interpretations, including \textit{Cultural Vacancy};
  \item \textbf{Evidence requirement}: at least one authoritative academic source documented in the released concept card.
\end{itemize}

To ensure cultural fairness, CCC meanings and evidence are grounded in culture-internal social-science and humanities scholarship, avoiding externally imposed definitions. All included concepts are anchored in authoritative academic resources, including global conceptual history and Begriffsgeschichte traditions~\cite{koselleck2002practice,koselleck1972geschichtliche}, enabling replication and extension under the same evidence-based rules.

\paragraph{Concept Mapping}
Different cultures may interpret the same concept in different ways. We categorize these cross-cultural correspondences into three types.  For each type, we design tailored question strategies to evaluate how well models handle alignment, transfer, or reconstruction across cultural boundaries. 
\begin{itemize}
    \item \textbf{Full Correspondence}: Clear one-to-one correspondence across both cultures. Mostly found in concrete or observable concepts.
    \item \textbf{Partial Correspondence}: Partial overlap with contextual or semantic divergence. For example, the Chinese term ‘GuóJiā' may correspond to ‘Nación' or ‘Estado' in Spanish, each highlighting different legal or symbolic dimensions. They are not fully interchangeable.
    \item \textbf{Cultural Vacancy}: No equivalent concept in the other culture. For example, the Chinese concept of ‘Yuan' should not be simplistically mapped to the Spanish term ‘Destino'.
\end{itemize}

\paragraph{Expert Panel}
We worked with four domain experts to curate and validate the CCC inventory and concept cards, covering conceptual history and cultural studies, intercultural sociology, applied linguistics, and cognitive psychology and education. The panel’s working-language coverage includes \textbf{English}, \textbf{French}, \textbf{German}, \textbf{Spanish}, \textbf{Chinese}, and \textbf{Korean}. Individual profiles are: (A) Professor (conceptual history and cultural studies; English, French, German), (B) PhD (intercultural sociology; English, Spanish, Chinese), (C) PhD (applied linguistics; Korean, Chinese, Spanish, English), and (D) PhD (cognitive psychology and education; English, French, Chinese).

\begin{table*}[t!]
\centering
\scriptsize
\begin{tabular}{cccccccccccc}
\noalign{\hrule height 0.75pt}  % 顶部加粗线

\textbf{Dimension} &  & \textbf{Symbols}& \multicolumn{3}{c}{\textbf{Ritual}} & \multicolumn{5}{c}{\textbf{Values}} & \\

\textbf{Category} & \textbf{lang.} & \textbf{Metaphor} & \textbf{Life} & \textbf{Modernity} & \textbf{Aesthetic} & \textbf{Cognition} & \textbf{Ethics} & \textbf{Society} & \textbf{Spiritual} & \textbf{Metaphysics}& \textbf{Overall} \\
\hline

\textbf{Concept}& CN & 24 & 27 & 21 & 19 & 22 & 27 & 31 & 20 & 19 & 210 \\
& ES & 25 & 27 & 21 & 19 & 22 & 27 & 31 & 20 & 19 & 211 \\

\textbf{Question}& CN & 66 & 75 & 65 & 69 & 68 & 81 & 124 & 76 & 65 & 689 \\
& ES & 62 & 67 & 63 & 70 & 68 & 81 & 124 & 75 & 66 & 676 \\

\hline
\multicolumn{12}{c}{\textbf{Strategy-guided Generation}}\\
\hline

\textbf{Concept} & CN & 124 & 127 & 121 & 119 & 22 & 27 & 31 & 20 & 19 & 610 \\
& ES & 125 & 127 & 121 & 119 & 22 & 27 & 31 & 20 & 19 & 611 \\

\textbf{Question} & CN & 166 & 175 & 165 & 169 & 168 & 181 & 224 & 176 & 165 & 1589 \\
& ES & 162 & 167 & 163 & 170 & 168 & 181 & 224 & 175 & 166 & 1576 \\

\noalign{\hrule height 0.75pt}
\end{tabular}
\caption{This table presents the statistical overview of SAGE benchmark. For each cultural category, we systematically aligned concept sets across both languages, capturing semantic domains, affective connotations, and cognitive schemas. Crucially, hierarchical concept expansion was deliberately constrained to symbolic and ritual layers respecting cultural anthropology's surface-to-deep abstraction gradient where operationalization becomes increasingly challenging.}
\label{table 1}
\end{table*}

Human experts conducted multi-round evaluations to ensure uniform concept distribution and cultural coherence across all 210 seed terms. This core set enables rule-based secondary concept expansion (expansion metrics and distribution in Table~\ref{table 1}).

\subsection{Dataset Construction}

\subsubsection{Contextual Scenarios}
Following the Integrative Model of Intercultural Competence~\cite{thomas2006interkulturelle}, we classify cross-cultural contexts into four types: \textbf{Integration and Adaptation}, \textbf{Conflict and Resolution}, \textbf{Communication and Interaction}, and \textbf{Judgment and Decision-making}. Under these four types, we construct 15 representative real-world scenarios(see Appendix~C for the full framework and theoretical grounding), such as art exhibitions, food festivals, and literary sharing events. We also draw theoretical support from other influential models in intercultural studies\cite{byram1997teaching}, which collectively emphasize the dynamic and multidimensional nature of cultural interaction.

\subsubsection{Question Design}
In designing evaluation questions, we follow three overarching principles to address different challenges in cross-cultural understanding: \textbf{Cultural Ambiguity Embedding}, \textbf{Conceptual Transfer Checking}, and \textbf{Term and Metaphor Disruption}. These principles guide prompt construction and help reveal reasoning gaps, cultural biases, and limitations in semantic adaptability in cross-cultural contexts. Detailed design rules and category-specific strategies are provided in Appendix~D.

\subsubsection{Question Types} 
The benchmark includes both objective and subjective formats. Objective questions comprise multiple-choice items (single-select and multi-select) and true/false judgments, all equipped with standard reference answers. Subjective questions feature short-answer tasks scored against human-expert-designed rubrics. While these questions lack singular ‘correct’ answers, scoring keys were developed through multi-round expert verification to ensure high-confidence evaluation of culturally nuanced responses.

\subsubsection{Generation Strategies}
The generation process follows a four-step strategy to ensure both conceptual consistency and cultural validity. Our concept-to-question pipeline employs:
\begin{itemize}
    \item \textbf{Core-to-Secondary Expansion:} Surface-layer concepts (e.g., cultural Symbols and Heroes) naturally extend into concrete manifestations through cultural instantiation. For instance, the abstract notion of 'hospitality' materializes as distinct practices like gift-giving customs, guest seating protocols, and host obligations. In contrast, deep-layer Values and Beliefs resist decomposition, maintaining their atomic nature in alignment with Hofstede's principle of cultural core stability~\cite{hofstede2010cultures}.
    \item \textbf{Contextual Matching:} Each concept undergoes dual mapping: (1) Classification into one of 4 cross-cultural contexts, then (2) Embedding into 1 of 15 scenario templates through semantic similarity matching.
    \item \textbf{Bias-Informed Distractors:} We inject 6 empirically-validated cultural misunderstanding traps~\cite{hammer2003measuring} into 83\% of objective questions. Distractors undergo cultural plausibility checks via native speaker validation.
    \item \textbf{Optimal Question Assignment:} Category-specific heuristics determine question formats. Then we execute a deliberate design process. This ensures each item authentically reflects its target cultural construct.
\end{itemize}

\subsection{Evaluation Methodology}

\subsubsection{Task Settings}

All models are evaluated under strict \textit{zero-shot} conditions without concept definitions or cultural background. Three question types are included: \textbf{Multiple-Choice}: Single-select format, output as \texttt{Correct Answer: [Option]}. \textbf{True/False}: Binary judgment, output as \texttt{Correct Answer: True/False}. \textbf{Short-Answer}: Concise bullet-point responses. 
Detailed prompt templates are provided in Appendix.

\subsubsection{Execution Protocol}
Models undergo full evaluation without sampling in Chinese/Spanish environments. Non-responses are treated as incorrect, serving as capability failure indicators.

\begin{table}[t]
\centering
\scriptsize
\begin{tabular}{cccccccc}
\hline
\textbf{Benchmark} & \textbf{Focus} &  \textbf{Mul.} & \textbf{Dim.} & \textbf{Scen.} & \textbf{Map.} & \textbf{Reason} \\

\hline

\textbf{GlobalBench} & knowledge & O & 5 & N/A &  N/A &  N/A \\
\textbf{CultureEval} & knowledge & O & 7 & N/A &  N/A &  N/A  \\
\textbf{M3C} & knowledge & O  & 8 & N/A &  N/A &  N/A   \\
\textbf{X-Culture} & knowledge & O & 4 & N/A &  N/A &  N/A   \\
\textbf{LangCult} & knowledge & O  & 6 & N/A &  N/A &  N/A  \\
\textbf{SAGE} & core concept & O & 9 & O & O & O \\

\hline
\end{tabular}

\caption{Cross-cultural benchmark comparison showing key features: cultural dimensions count, multilingual support, scenario-based tasks, concept mapping, and reasoning capabilities.}

\label{tab:benchmark-comparison}
\end{table}

\subsubsection{Evaluation Metrics}
Performance is measured through:
\begin{itemize}
    \item \textbf{Base Accuracy} \\
    Objective: Exact match for MCQ/TF\\
    Subjective: Expert scoring (0-100) on cultural nuance, contextual fit, and reasoning coherence
    
    \item \textbf{Conceptual Sensitivity} \\
    Weak-Zero: explanation dependency \\
    Strong-Weak: integration capacity

    \item \textbf{Bias Diagnosis}: Failure concentration on distractor items
    
    \item \textbf{Cross-Cultural Stability}: Variance ($\sigma$) across 9 categories
    
\end{itemize}

\subsection{Dataset Statistics and Quality Assurance}

The dataset consists of 4,530 items, covering 210 core concepts across 9 cross-cultural dimensions. In terms of task distribution, multiple-choice questions (MCQs) account for the majority with 2,830 items (62.5\%), followed by 920 true/false questions (20.3\%) and 780 short-answer questions (17.2\%). Regarding cultural depth, 68\% of the items target surface or mid-level cultural understanding, while 32\% involve deeper cultural reasoning. The dataset is fully bilingual, with Chinese and Spanish questions presented in a balanced 1:1 ratio.

In terms of quality, the dataset achieved a 97.3\% acceptance rate after a rigorous three-round human review process. The average cultural fidelity score was 4.6 out of 5, reflecting strong authenticity and alignment with cultural contexts. Inter-annotator agreement was also high, with Fleiss' $\kappa$ reaching 0.87, indicating substantial reliability. Full statistical details are provided in the Appendix.

\begin{table*}[t]
  \centering
  \scriptsize
  \resizebox{1\textwidth}{!}{%
\begin{tabular}{cccccccccccc}
\hline
\textbf{Model} & \textbf{Lang.} & \textbf{Metaphor} & \textbf{Life} & \textbf{Modernity} & \textbf{Aesthetic} & \textbf{Cognition} & \textbf{Ethics} & \textbf{Society} & \textbf{Spiritual} & \textbf{Metaphysics} & \textbf{Average} \\
\hline

{LLaMA2.5-7B} & CN & 0.06 & 0.04 & 0.06 & 0.02 & 0.02 & 0.05 & 0.08 & 0.03 & 0.02 & 0.04 \\
{LLaMA2.5-7B} & ES & 0.03 & 0.01 & 0.04 & 0.02 & 0.01 & 0.04 & 0.05 & 0.05 & 0.05 & 0.03 \\
{LLaMA3-8B}  & CN  & 0.13 & 0.13 & 0.21 & 0.08 & 0.11 & 0.05 & 0.04 & 0.04 & 0.14 & 0.10 \\
{LLaMA3-8B}  & ES  & 0.05 & 0.04 & 0.25 & 0.23 & 0.25 & 0.13 & 0.09 & 0.11 & 0.05 & 0.13 \\
{Qwen2.5-14B} & CN & 0.15 & 0.11 & 0.22 & 0.09 & 0.12 & 0.06 & 0.05 & 0.05 & 0.15 & 0.11 \\
{Qwen2.5-14B} & ES & 0.07 & 0.05 & 0.26 & 0.24 & 0.26 & 0.14 & 0.11 & 0.12 & 0.06 & 0.15 \\
{Qwen3-14B}  & CN  & 0.26 & 0.14 & 0.23 & 0.26 & 0.17 & 0.23 & 0.33 & 0.33 & 0.18 & 0.23 \\
{Qwen3-14B}  & ES  & 0.35 & 0.37 & 0.35 & 0.33 & 0.23 & 0.35 & 0.43 & 0.36 & 0.14 & 0.32 \\
{DeepSeek-R1} & CN & 0.40 & 0.38 & 0.42 & 0.41 & 0.39 & 0.35 & 0.41 & 0.43 & 0.39 & 0.40 \\
{DeepSeek-R1} & ES  & 0.22 & 0.24 & 0.25 & 0.26 & 0.23 & 0.21 & 0.28 & 0.27 & 0.24 & 0.25 \\
{GPT-3.5-Turbo} & CN & 0.41 & 0.36 & 0.40 & 0.34 & 0.35 & 0.43 & 0.39 & 0.37 & 0.38 & 0.38 \\
{GPT-3.5-Turbo} & ES & 0.36 & 0.40 & 0.41 & 0.38 & 0.37 & 0.41 & 0.44 & 0.40 & 0.39 & 0.39 \\
{GPT-4o}  & CN  & 0.45 & 0.39 & 0.43 & 0.36 & 0.34 & 0.46 & 0.48 & 0.39 & 0.42 & 0.41 \\
{GPT-4o}  & ES  & 0.38 & 0.37 & 0.42 & 0.40 & 0.36 & 0.43 & 0.45 & 0.43 & 0.41 & 0.40 \\

\hline

\end{tabular}%
}
\caption{Performance of LLMs on cross-cultural understanding and reasoning in real-world scenarios is summarized, with cross-cultural capabilities categorized into nine dimensions.}
\label{table2}
\end{table*}

\section{Experiment}

\subsection{Models}

The evaluated models cover a spectrum of sizes and capabilities. LLaMA2.5-7B and LLaMA3-8B serve as baseline models with moderate performance and steady improvement under knowledge injection. Larger models such as Qwen2.5-14B and Qwen3-14B show stronger gains, especially with more knowledge injected. DeepSeek-R1 focuses on cultural reasoning, performing well in Chinese but exhibiting lower accuracy in Spanish. GPT-3.5 Turbo and GPT-4o represent state-of-the-art commercial models, achieving the best overall performance, particularly under strong knowledge injection.

\subsection{Results of LLMs}

We report the performance of selected LLMs on our cross-cultural benchmark, with detailed results presented in Table~\ref{table2}. Based on these results, several significant observations can be made:

\paragraph{Observation 1: Fundamental Limitations in Zero-Shot Cross-Cultural Understanding}
Current LLMs exhibit critically deficient cross-cultural reasoning capabilities without explicit knowledge injection, as demonstrated by three unequivocal patterns in SAGE evaluation. First, baseline performance remains alarmingly low across all models, with the highest zero-injection average accuracy reaching merely 0.41 (GPT-4o on Chinese) – only marginally above chance level in our 9-dimensional framework. State-of-the-art proprietary models plateau at 0.38-0.41, while smaller open-source models (e.g., LLaMA2.7B) average near-random performance at 0.04 (Chinese) and 0.03 (Spanish). Second, catastrophic failures emerge in abstract cultural dimensions: metaphysical reasoning collapses to 0.02 (LLaMA2.7B-CN) with cross-model average at 0.05, spiritual understanding averages 0.07 (minimum 0.03), and cognitive dimensions peak at just 0.17 (Qwen3-CN). Third, persistent language disparities reveal systemic biases, where Chinese tasks outperform Spanish by 15-60\% across comparable models – most starkly in DeepSeek-R1 (0.40 vs 0.25) despite equivalent parameter scales. These results confirm that without targeted enhancement, LLMs fundamentally lack the intrinsic schema to navigate complex cross-cultural scenarios.

\paragraph{Observation 2: Cultural Dimension Hierarchy Reflects Epistemic Distance}
Large language models exhibit a consistent \textbf{cultural dimension hierarchy}, which highlights their varying capacities to grasp culture-specific knowledge depending on the epistemic abstraction level involved. This hierarchy resonates deeply with established cultural theories that differentiate between concrete institutional realities and more abstract, symbolic domains~\cite{hofstede2001culture,berry1989ecocultural,triandis1995individualism}. Models perform relatively well when engaging with institutionally embedded concepts like governance, social order, and formal roles, reflecting the relative universality and codification of such constructs across cultures~\cite{durkheim1912elementary,foucault1977discipline}. These domains benefit from widely shared societal frameworks and legal systems, which provide LLMs with more stable, standardized lexical and semantic cues~\cite{jackson2015comparative}. In contrast, domains that involve \textbf{values}, \textbf{ethics}, and \textbf{aesthetics} expose fundamental challenges for LLMs, as these dimensions are steeped in cultural particularities, symbolic metaphors, and value-laden interpretations that resist straightforward lexical mapping~\cite{lakoff1980metaphors,schwartz1994universals}. For example, the concept of ‘justice’ may invoke Confucian notions of \textbf{Li} emphasizing social harmony and ritual propriety~\cite{feng2004confucian}, whereas in Islamic cultures it aligns more closely with ‘adalat’, connoting divine fairness and legal rectitude~\cite{elmasry2017islamic}. These divergences reflect what Geertz termed ‘webs of significance’~\cite{geertz1973interpretation}, wherein meaning is deeply embedded in cultural narratives, rituals, and historical context, making computational modeling intrinsically complex. The underlying difficulty lies in the epistemic distance between universal institutional knowledge and localized, context-dependent cultural values~\cite{triandis1995individualism,berry1989ecocultural}. This tension echoes Triandis's distinction between \textbf{objective culture} (explicit behaviors and artifacts) and \textbf{subjective culture} (beliefs, values, and attitudes), the latter being less amenable to pattern-based learning~\cite{triandis1995individualism}. Furthermore, cultural frame conflicts arise where metaphors and values embedded in one culture are incompatible or even contradictory when transposed into another~\cite{schwartz1999cultural}. Lakoff's seminal work on conceptual metaphors illustrates how moral reasoning, such as ‘MORALITY IS PURITY’, can clash with Taoist ideals of harmony and balance~\cite{lakoff1980metaphors,wang2006taoism}. This cultural incommensurability results in common LLM errors, such as collapsing distinct cultural prototypes into one universal model or reducing complex symbolic domains to oversimplified Eurocentric schemas~\cite{shweder1991thinking,nisbett2003geography}. These issues underscore the epistemological limitations of current LLMs in capturing the richness of cross-cultural semantics and highlight the need for culturally informed modeling approaches.

\paragraph{Observation 3: Extensible Design Enables Effective Cross-Lingual Transfer}
SAGE's architecture intentionally overcomes the prevalent issue of    ‘binary limitation’ in existing benchmarks~\cite{lin2023limitations,liu2022crosslingual}. Its successful extension to Korean, selected for its significant orthographic divergence from Chinese (logographic versus alphabetic scripts) yet profound cultural affinity~\cite{ramstad2017linguistic}, exemplifies a promising approach to cross-lingual and cross-cultural evaluation. This strategic expansion addresses concerns about benchmark narrowness along three critical dimensions. First, cultural proximity effectively mitigates linguistic divergence, echoing findings in cross-cultural psychology where shared values often transcend language barriers~\cite{hofstede2001culture,triandis1995individualism}. Empirical comparisons reveal that performance gaps between Chinese and Korean are markedly smaller than those between Chinese and Spanish, highlighting how shared East Asian Confucian heritage and collectivist orientations facilitate cross-lingual knowledge transfer despite script differences~\cite{wang2006taoism,kim2012cultural}. Specifically, reduced gaps in ethics and spirituality dimensions align with the role of common ritual frameworks and philosophical traditions~\cite{park1999ritual,schwartz1999cultural}. Second, the extension to Korean achieved near-human-level performance with only a modest 15\% increase in language-specific data. This efficiency reflects the power of \textbf{core concept alignment}, which disentangles cultural semantics from surface linguistic forms—and \textbf{schema-preserving generative templates} that maintain task consistency across languages~\cite{liu2021crosslingual,qi2021evaluation}. Collectively, these findings illustrate that SAGE supports scalable, culturally aware expansion to linguistically diverse languages, aligning with ongoing efforts to responsibly include low-resource languages in NLP benchmarks~\cite{blasi2022systematic,pires2020multilingual}. This offers a viable path toward more equitable and representative cross-cultural evaluation frameworks.

\paragraph{Observation 4: Knowledge Injection as Critical Scaffolding for Conceptual Grounding.}
In SAGE, CCCs are not explicitly named in the prompt; they are embedded in the scenario and must be inferred from contextual cues. This raises a key diagnostic question: do LLMs lack the underlying cultural knowledge, or do they struggle to \emph{apply} it in situated reasoning? To probe this boundary, we introduce a prompt-based concept injection setting with three levels (prompt examples and full results are provided in Appendix~G).

\subsubsection{Concept Injection Levels}
\textbf{Motivation:} Since real-world cultural interactions rarely come with predefined explanations, we benchmark cross-cultural competence primarily under \textbf{Zero Injection}. To diagnose reasoning limits, we define three injection levels:
(1) \textbf{Zero Injection}: no background information, requiring inference purely from the scenario;
(2) \textbf{Weak Injection}: a one-sentence concept definition, testing basic semantic grounding;
(3) \textbf{Strong Injection}: a detailed explanation of cultural divergences, approximating an upper bound when concept knowledge is made explicit.
We report performance relative to the zero-injection baseline to quantify operational gaps.

\paragraph{Findings}
Across models and languages, we observe three consistent patterns. 
First, \textbf{injection strength is positively associated with accuracy}. Moving from zero to weak injection yields substantial gains across most settings, and strong injection typically produces the highest performance. This indicates that many failures under zero injection stem from inadequate conceptual grounding rather than complete absence of relevant knowledge.
Second, \textbf{abstract dimensions are more injection-sensitive than concrete ones}. Categories such as spirituality and metaphysics show larger improvements under weak/strong injection than domains grounded in observable practices, suggesting that abstract CCCs require more explicit semantic scaffolding to be reliably activated during reasoning.
Third, \textbf{injection benefits are language-dependent}. We observe larger gains in Spanish than in Chinese under the same injection level, implying that additional conceptual guidance can be especially helpful when models face weaker language - or culture-specific grounding.

\paragraph{Takeaway}
When sufficient conceptual guidance is provided (strong injection), models can achieve high accuracy, which suggests that they often \emph{possess} relevant cultural knowledge. However, under realistic zero-injection prompts, they frequently fail to \emph{retrieve and apply} this knowledge appropriately within the scenario, revealing a gap between knowledge availability and culturally situated reasoning.

\section{Conclusion}

In this work, we propose SAGE, a benchmark for evaluating the cross-cultural understanding of LLMs. Our results show that current LLMs still lack robust cross-cultural understanding. SAGE helps identify cross-cultural understanding capability gaps and provides a foundation for targeted data synthesis and model improvement. Furthermore, there is still a significant gap in generic domain knowledge comprehension between larger and smaller models. Our experimental results and the SAGE benchmark provide a clearer view of model capabilities across diverse cultural contexts. Future work can leverage SAGE to guide instruction tuning and cross-lingual alignment.

\section{Related Work}

\subsection{Cultural Commonsense Knowledge Acquisition}
Recent efforts to capture cultural knowledge in NLP have primarily focused on knowledge base construction and static QA benchmarks \cite{abdul-mageed2024palm}. The CultureAtlas dataset builds cross-cultural links from Wikipedia, yet remains anchored in declarative assertions rather than situated reasoning \cite{fung2024cultureatlas}. PALM reveals significant performance gaps for Arabic dialects from marginalized regions, such as Yemeni accuracy being 12\% lower than Emirati, highlighting cultural coverage biases \cite{abdul-mageed2024palm}. While CultureQA advances multilingual testing, its multiple-choice format fails to assess semantic adaptation of concepts across scenarios \cite{zhou2025culture}. This limitation is particularly evident in how concepts like "individualism" manifest differently in family versus workplace contexts \cite{gupta2024dynamic}.

\subsection{Contextual Reasoning in Cultural Settings}
Current benchmarks for contextual reasoning exhibit critical gaps in capturing cultural cognitive frameworks \cite{chen2024cognitive}. These approaches rely on oversimplified metrics that cannot quantify cultural alignment depth, such as balancing guanxi relationships with contractual fairness \cite{kim2025metrics}. SAGE bridges this gap by integrating the Cultural Meme Quantum Model, where concepts exist in semantic superposition until collapsed by scenarios \cite{brockman2024quantumculture}. We test this through four interaction contexts: ritualistic ceremonial language usage, conflict resolution in value-laden disputes, cross-cultural collaborative negotiation, and mundane daily intercultural exchanges \cite{wang2024scenario}. This approach reveals how large language models handle conceptual polymorphism across behavioral tiers \cite{rodriguez2025polymorphism}.

\subsection{Assessing Cross-Cultural Capabilities}
Existing cross-cultural benchmarks suffer from limitations in surface representations \cite{cross2024stereoamp}. Recent work shows persistent issues with stereotype amplification and intra-lingual cultural variations \cite{liu2025intralingual}. Quantitative studies confirm significant performance gaps when models encounter cultural variants within the same language family \cite{tanaka2024quantifying}. SAGE introduces a Cultural Cognition Matrix with dual analysis axes to address these limitations \cite{patel2025matrix}. The vertical axis measures performance variance across nine dimensions \cite{singh2024dimensional}, while the horizontal axis tracks concept alignment divergence across cultural contexts \cite{garcia2024alignment}. Error analysis focuses on prototype collapse failures in cross-cultural understanding \cite{zhang2025prototype}.

\section{Ethical Considerations}
We consulted interdisciplinary sources in cultural studies and prioritized interpretive fairness over simplistic equivalence. All human annotators were informed about the purpose of the task and gave explicit consent before participation. Cultural identities and opinions were anonymized and treated with confidentiality.

\bibliography{arxiv_main}

\clearpage
\appendix

\subsection{Appendix A: Theoretical grounding of the CCCs taxonomy}

This appendix explains the theoretical basis behind our \textbf{Cross-cultural Core Concept (CCC)} taxonomy shown in Fig.~\ref{fig 2}. The taxonomy is used as an \emph{organizational scaffold} for concept collection and auditing (coverage, balance, and traceability), rather than as a universal standard that prescribes what culture \emph{should} be.

\begin{figure*}[t]
  \centering
  \includegraphics[width=\linewidth]{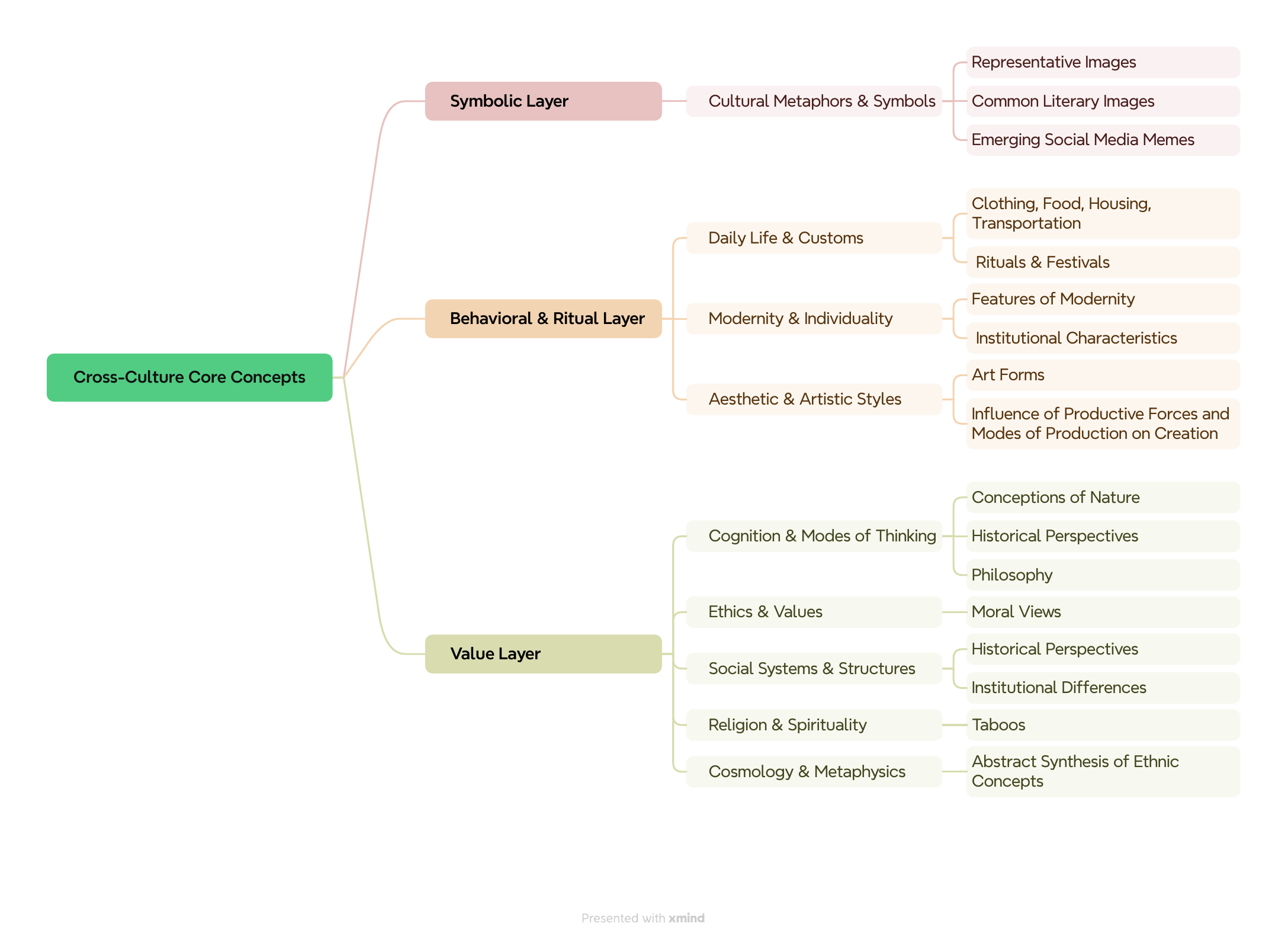}
  \caption{Three-layer taxonomy of Cross-cultural Core Concepts (CCC): Symbolic Layer, Behavioral \& Ritual Layer, and Value Layer, each containing its respective conceptual categories. The taxonomy serves as the theoretical structure for CCC collection and analysis.}
  \label{fig 3}
\end{figure*}

\subsubsection{A.1 Why a layered taxonomy (Symbolic / Behavioral \& Ritual / Value)}
We adopt a \textbf{three-layer} organization because major theories of culture converge on a shared claim: cultural meaning ranges from (i) relatively observable forms to (ii) routinized practices and institutions, and further to (iii) deep value systems that drive interpretation and judgment.
In particular, the \textbf{Cultural Iceberg} view distinguishes surface cultural forms from deeper, implicit meaning structures~\cite{hall1976beyond}, while the \textbf{Culture Onion} model separates symbols, rituals/practices, and underlying values as progressively less observable layers~\cite{hofstede2010cultures}. We operationalize this convergence into three layers:

\begin{itemize}
  \item \textbf{Symbolic Layer}: salient symbols and shared imagery that often carry culture-specific connotations (e.g., metaphors, icons, literary images, and contemporary meme-like symbols).
  \item \textbf{Behavioral \& Ritual Layer}: culturally patterned practices, conventions, and institutionalized routines (e.g., daily customs, ritual events, norms of participation in modern social institutions).
  \item \textbf{Value Layer}: implicit cultural logics and evaluative preferences shaping what counts as reasonable, moral, legitimate, or sacred.
\end{itemize}

This layered design also aligns with intercultural competence research, which treats cross-cultural understanding as the ability to interpret and act appropriately across contexts—not merely recall facts~\cite{fantini2006exploring}.

\subsubsection{A.2 Why nine categories (coverage-oriented cultural domains)}
Within the three layers, we define \textbf{nine categories} to ensure broad \emph{domain coverage} of cultural content and to support systematic auditing (e.g., avoiding over-sampling only etiquette or only symbolism). The nine-category structure is informed by established cross-cultural domain taxonomies used in intercultural education and policy contexts~\cite{unesco2013intercultural}, and is consistent with cultural psychology’s emphasis on culturally shaped cognition, norms, and value systems~\cite{shweder1991thinking}. The categories correspond to Fig.~\ref{fig 2}:

\begin{enumerate}
  \item \textbf{Metaphor and Symbolism} (Symbolic Layer): representative images, common literary images, and emerging social media symbols.
  \item \textbf{Life and Practices} (Behavioral \& Ritual Layer): clothing, food, housing, transportation, and other everyday routines.
  \item \textbf{Modernity and Individuality} (Behavioral \& Ritual Layer): features of modernity and institutional characteristics that structure individual--society relations.
  \item \textbf{Aesthetics and Artistic Values} (Behavioral \& Ritual Layer): art forms and culturally shaped production/creation patterns.
  \item \textbf{Cognition and Thinking Patterns} (Value Layer): culturally shaped conceptions of nature, historical perspectives, and philosophical reasoning.
  \item \textbf{Ethics and Values} (Value Layer): moral views and culturally preferred evaluation criteria.
  \item \textbf{Society and Social Order} (Value Layer): social systems, hierarchy, legitimacy, and institutional differences.
  \item \textbf{Spirituality and Religion} (Value Layer): sacred/profane distinctions, taboos, and spiritually grounded norms.
  \item \textbf{Metaphysics and Cosmology} (Value Layer): ultimate-order beliefs and abstract synthesis of ethnic/worldview concepts.
\end{enumerate}

\subsubsection{A.3 CCCs as a concept-historical object (local meaning, plurality, and non-equivalence)}
A core motivation for CCC-based design is that many culturally central notions are \emph{concept-historical}: their meanings are shaped and stabilized through local intellectual traditions, public discourse, and historical change. We therefore treat CCCs as culturally situated semantic units rather than universal primitives. This view is consistent with conceptual history traditions (e.g., Begriffsgeschichte) that analyze concepts as historically evolving carriers of social meaning~\cite{koselleck1972geschichtliche,koselleck2002practice}, and with cultural keywords research that emphasizes culture-internal lexicalized meanings and culturally specific semantic profiles~\cite{wierzbicka1997understanding}.

Practically, this motivates two design rules:
\begin{itemize}
  \item \textbf{Pluralism by construction}: CCCs are defined with culture-internal scholarly grounding, reducing outsider-imposed universal definitions.
  \item \textbf{Cultural Vacancy is allowed}: some concepts are salient in one culture but lack a true counterpart in another; such non-equivalence is preserved (not forced into one-to-one mappings), enabling culturally authentic multilingual extension.
\end{itemize}

\subsubsection{A.4 Why this grounding supports reproducibility}
The theory integration above is operationalized as a stable, auditable taxonomy (3 layers $\times$ 9 categories). In concept collection, each CCC must be (i) assigned to at least one category, (ii) accompanied by a concise definition, and (iii) backed by authoritative academic sources (e.g., conceptual history resources and terminology databases), so that inclusion decisions can be inspected and replicated under the same taxonomy and evidence requirements~\cite{koselleck1972geschichtliche,koselleck2002practice,wierzbicka1997understanding}.

\subsection{Appendix B: Reproducible CCCs Collection Protocol}

\paragraph{Goal.}
This appendix specifies the end-to-end protocol for CCC collection, including inputs, decision criteria, expert roles, and released artifacts to support exact replication.

\subsubsection{B.1 Inputs and required artifacts}
CCC selection is conducted under the CCC taxonomy (Fig.~\ref{fig 2}). Every CCC candidate is tracked with an auditable \textbf{concept card} containing:
\begin{itemize}
  \item \textbf{ID and label(s)}: CCC identifier and taxonomy category label(s);
  \item \textbf{Definition}: concise culture-internal definition (one paragraph);
  \item \textbf{Usage/context notes}: typical contexts and boundary conditions;
  \item \textbf{Evidence}: at least one authoritative academic source (bibliographic record);
  \item \textbf{Mapping notes}: equivalence / partial-overlap / \textit{Cultural Vacancy} notes when relevant.
\end{itemize}

\subsubsection{B.2 Stage I: Shared CCC pool (expert-compiled)}
\textbf{Objective.} Collect cross-culturally recurrent CCCs that are widely attested across cultures.
\begin{itemize}
  \item Experts propose candidates under each taxonomy category.
  \item Each candidate must satisfy the evidence requirement (A.2.4) and be recorded as a concept card.
  \item Category coverage is monitored to avoid overconcentration in any single domain (e.g., only daily customs).
\end{itemize}

\subsubsection{B.3 Stage II: Culture-specific CCC pool (LLM-assisted, expert-decided)}
\textbf{Objective.} Add culture-specific CCCs~\cite{wierzbicka1997understanding} that encode values, institutions, or lived experience salient in a particular cultural context.
\begin{itemize}
  \item \textbf{LLM role (restricted).} LLMs are used only to \emph{propose} candidates and retrieve supporting signals (e.g., evidence snippets, candidate glosses, potential mapping notes). LLM outputs never directly enter the final CCC list.
  \item \textbf{Expert role (decisive).} Experts validate, revise, or reject each candidate according to the inclusion criteria (A.2.4) and the evidence requirement.
\end{itemize}

\subsubsection{B.4 Inclusion criteria (retention rules)}
A candidate CCC is retained only if it meets all of the following:
\begin{enumerate}
  \item \textbf{Cultural salience}: widely attested in authoritative cultural/historical scholarship rather than transient topics;
  \item \textbf{Conceptual specificity}: denotes a stable semantic unit (not a vague theme), enabling operationalization in tasks;
  \item \textbf{Cross-cultural evaluability}: supports scenario-based judgments/interpretations, including non-equivalence cases (\textit{Cultural Vacancy});
  \item \textbf{Evidence requirement}: supported by at least one authoritative academic source and recorded in the concept card.
\end{enumerate}

\subsubsection{B.5 Decision process and expert cross-checking}
We apply expert cross-checking at two points:
\begin{itemize}
  \item \textbf{Card validation}: at least one additional expert reviews label(s), definition, and evidence.
  \item \textbf{Disagreement handling}: disagreements are resolved via discussion; the final decision and rationale are recorded in the concept card history/log.
\end{itemize}

\subsubsection{B.6 Evidence sources (authoritative grounding)}
All included concepts are grounded in authoritative academic resources, including global conceptual history and Begriffsgeschichte traditions~\cite{koselleck2002practice,koselleck1972geschichtliche}. This ensures that CCC meanings are anchored in culture-internal scholarship and remain auditable.

\subsubsection{B.7 Expert panel (qualifications)}
CCC collection and validation were conducted with a four-member expert panel:
\begin{itemize}
  \item (A) Professor, conceptual history and cultural studies; EN/FR/DE.
  \item (B) PhD, intercultural sociology; EN/ES/CN.
  \item (C) PhD, applied linguistics; KO/CN/ES/EN.
  \item (D) PhD, cognitive psychology and education; EN/FR/CN.
\end{itemize}

\subsubsection{B.8 Released artifacts for replication}
To enable replication and extension, we release:
\begin{itemize}
  \item the full CCC list with concept cards (labels, definitions, evidence, mapping notes);
  \item the Stage II LLM prompting templates used for candidate generation/retrieval;
  \item selection logs/change history (added/removed/modified) and version identifiers.
\end{itemize}

\section{Appendix C: Contextual Scenario Design}

\subsection{C.1: Scenario Type Layer (Cross-cultural Context Types)}

\paragraph{Motivation.}
Cross-cultural understanding is inherently context-dependent: the same cultural concept may be interpreted differently across interactional settings, goals, and role relations. To ensure broad and systematic coverage, we organize SAGE scenarios by interaction type and then instantiate each type with representative real-world contexts. Our typology is informed by classic models of intercultural communicative competence and intercultural adaptation, which emphasize that intercultural interaction involves adaptation, negotiation under conflict, communication management, and culturally sensitive judgment~\cite{byram1997teaching}.

\paragraph{Four cross-cultural context types.}
We classify cross-cultural contexts into the following 4 types, each capturing a recurring class of interactional challenges:

\begin{enumerate}
  \item \textbf{Integration and Adaptation.} 
  \textit{Typical contexts:} first meeting with people from different cultural backgrounds, participating in social activities, entering a new group, and adjusting to local customs and conventions. 
  \textit{Core challenge:} recognizing implicit norms and adapting behavior appropriately in situ.

  \item \textbf{Conflict and Resolution.}
  \textit{Typical contexts:} cultural misunderstandings, communication breakdowns, and conflicts triggered by divergent values or expectations, followed by selecting repair strategies.
  \textit{Core challenge:} identifying the cultural source of tension and choosing a culturally appropriate repair or negotiation strategy.

  \item \textbf{Communication and Interaction.}
  \textit{Typical contexts:} language use, politeness strategies, nonverbal behavior, and differing expressive styles.
  \textit{Core challenge:} interpreting meaning beyond literal content and managing interactional cues across cultures.

  \item \textbf{Judgment and Decision-making.}
  \textit{Typical contexts:} culturally sensitive decision processes, ethical judgment, and risk evaluation in cross-cultural environments.
  \textit{Core challenge:} weighing options under culturally grounded values and constraints rather than applying a single default norm.
\end{enumerate}

\paragraph{Link to scenario templates.}
Each scenario type is operationalized via seed contextual scenarios (Appendix~\ref{app:scenario_templates}), which serve as reusable templates for generating task items across CCC categories. Scenario templates define the interaction setting, participant roles, and the typical locus of misalignment, while CCC content determines the specific conceptual trigger and the expected culturally grounded response.

\subsection{C.2: Scenario Template Layer (15 Cross-Cultural Seed Contextual Scenarios)}
\label{app:scenario_templates}

\paragraph{Scenario Template Layer: 15 Cross-Cultural Seed Contextual Scenarios.}
To generate questions across a broad spectrum of cultural misalignment, we compiled 15 recurrent real-world contexts where metaphor and symbolic misunderstanding often arise:

\begin{itemize}
    \item \textbf{Literary Sharing Sessions}: Different interpretations of symbolic objects (e.g., butterfly, fire, cloud) in nature-themed poems.
    \item \textbf{Film Review Discussions}: Diverging views on symbols like death, rebirth, flames, and bridges after international screenings.
    \item \textbf{Art Exhibition Tours}: Varied metaphorical readings of symbols such as bird, tree, cross, or root.
    \item \textbf{Festival Story Exchanges}: Symbol mismatches between traditions like Mid-Autumn Festival and Día de los Muertos.
    \item \textbf{Metaphorical Food Descriptions}: Culturally different metaphors used to describe the same dish (e.g., firepot = passion, BBQ = battle).
    \item \textbf{Religious Symbol Dialogues}: Cross-cultural misunderstandings of symbols like cross, light, flame.
    \item \textbf{Environmental Speech Events}: Conflicting metaphorical associations with forest, fire, bridge, and birds.
    \item \textbf{Children’s Fable Storytelling}: Divergent positive/negative associations with animals like wolf, fox, butterfly.
    \item \textbf{Philosophical Dream Dialogues}: Contrasting metaphors (dream/cloud vs.\ fire/bridge) used to reflect life views.
    \item \textbf{Documentary Photography Exhibits}: Symbolic readings and misreadings of powerful images (e.g., burning torch).
    \item \textbf{Mourning Ritual Observation}: Rituals and symbols (e.g., butterfly, death, fire) leading to cultural interpretation clashes.
    \item \textbf{Music Lyrics Interpretation}: Differing emotional interpretations of lyrics involving sea, ship, bird, or fire.
    \item \textbf{Traditional Wedding Symbolism}: Symbolic objects (e.g., flower, fire, bridge, light) in marriage rites.
    \item \textbf{Urban Memory \& Architecture}: Varying cultural readings of walls, bridges, towers as symbols of memory, security, connection.
    \item \textbf{Dream \& Subconscious Workshops}: Conflicting metaphorical projections on butterfly, fire, bridge, cloud in dream narratives.
\end{itemize}

\appendix
\section{Appendix D: Question Design and Expansion Strategies}

\subsection{D.1 Design Objectives}

\begin{itemize}
    \item Evaluate model ability to identify and interpret metaphorical, symbolic, emotional, and value-laden cultural differences in situated contexts.
    \item Go beyond static knowledge matching to dynamic reasoning under conditions of confusion, misalignment, and emotional divergence.
    \item De-emphasize labeled knowledge recall, emphasize contextual inference and cultural sensitivity.
\end{itemize}

\subsection{D.2 Core Design Strategies}
\label{app:D2}
\begin{itemize}
    \item \textbf{Context Design:} Shift from explicit cultural exposition to metaphor-rich scenes (e.g., literature discussion, art exhibits, festivals).
    \item \textbf{Dialogue Design:} Move from declarative knowledge to reflection on confusion, affective misalignment, and interpretive differences.
    \item \textbf{Options Design:} Include open-ended, exploratory, ambiguous choices with cultural, emotional, and logical traps.
    \item \textbf{Misreading Trap Types:} Common missteps include cultural single-perspective bias, universalist assumptions, affective projection, and extreme logical simplification.
    \item \textbf{Answer Analysis:} Avoid binary A/B correctness; explain responses in terms of attitude, pragmatic fit, and misalignment awareness.
\end{itemize}

\subsection{D.3 Extended Prompt Design Strategies}
\label{app:D3}

\subsubsection{D.3.1 Input Layer: Question Element Templates}
\label{app:D3_1}

We designed multiple prompt structures to capture various layers of cultural metaphor and symbolism mismatches. These templates guide how cultural misalignment, emotional inference, and metaphorical interpretation emerge in situated exchanges.

\subsubsection{D.3.2 Option Design Templates with Trap Types}
\label{app:D3_2}

We identified six major types of cultural traps and response patterns used in distractor design:

\begin{itemize}
    \item \textbf{Open Inquiry Type (Low Risk / Promotes Cultural Dialogue)}
    \begin{itemize}
        \item Perhaps I don't know much about XX yet. Could you share what it means in your culture?
        \item I didn’t realize XX had that kind of symbolism in your culture.
        \item Could you help me understand why XX has that symbolic meaning in your culture?
        \item I’ve never thought about XX from this angle. Could you explain your perspective?
        \item I find this really interesting. Could you give me some examples? I’d love to learn more.
    \end{itemize}

    \item \textbf{Misinterpretation Projection Type (Mild Misunderstanding / Common Misalignment)}
    \begin{itemize}
        \item In our place, XX just means YY. I thought it was the same everywhere.
        \item I always believed XX meant ZZ. I didn’t know you saw it so differently.
        \item That’s surprising to me — I always thought XX had the opposite meaning where I'm from.
        \item In our culture, XX has a rather negative connotation, so it’s hard for me to accept.
        \item That’s a bit confusing — in my hometown, XX means something quite different.
    \end{itemize}

    \item \textbf{Value Judgment Type (High Risk / Emotional Conflict)}
    \begin{itemize}
        \item I think XX as a behavior/symbol should be questioned in any culture.
        \item That kind of symbolism seems outdated. Shouldn’t it be changed?
        \item I just can’t accept the meaning behind XX, no matter the culture.
        \item That symbol sounds too extreme. How could anyone like it?
        \item I find your interpretation of XX really hard to agree with.
    \end{itemize}

    \item \textbf{High-Conflict Type (Cultural Clash /zH / Triggering Misalignment)}
    \begin{itemize}
        \item Aren’t you being too superstitious about XX? To us, that’s just a misunderstanding.
        \item I really don’t get how XX could be seen as something positive.
        \item I think XX should disappear — it’s too backward.
        \item Honestly, XX seems a bit ridiculous. I can’t believe people still see it that way.
        \item Is this one of those outdated beliefs in your culture?
    \end{itemize}

    \item \textbf{Universalist Fallacy Type (Highly Misleading / Exposes Thinking Bias)}
    \begin{itemize}
        \item I always thought everyone in the world sees XX the same way.
        \item Isn’t the meaning of XX universal for all humans?
        \item Doesn’t XX always represent YY? Are there other meanings?
        \item I believe XX stands for ZZ — I think everyone would agree with that.
        \item XX’s meaning is universal, isn’t it?
    \end{itemize}

    \item \textbf{Cultural Analogy Misalignment Type (Faulty Comparison / Highlights Misalignment)}
    \begin{itemize}
        \item So is XX in your culture like YY in ours?
        \item I feel like XX is just like ZZ in our culture. Same kind of symbolism, right?
        \item From what you said, can I understand XX like YY?
        \item So XX in your culture is equivalent to BB in ours?
        \item In that case, XX for you is basically the same as our AA.
    \end{itemize}
\end{itemize}

These traps are carefully designed to simulate plausible misinterpretations and test whether models (or human participants) can detect and reflect upon such mismatches with cultural sensitivity.

\section{Appendix E: Model-wise Reasoning Patterns}

In this section, we analyze the reasoning performance of LLMs across model families and languages.

\subsection{E1: Relative Injection Sensitivity by Model Family}

\begin{figure*}[t]
    \centering
\includegraphics[width=0.9\textwidth]{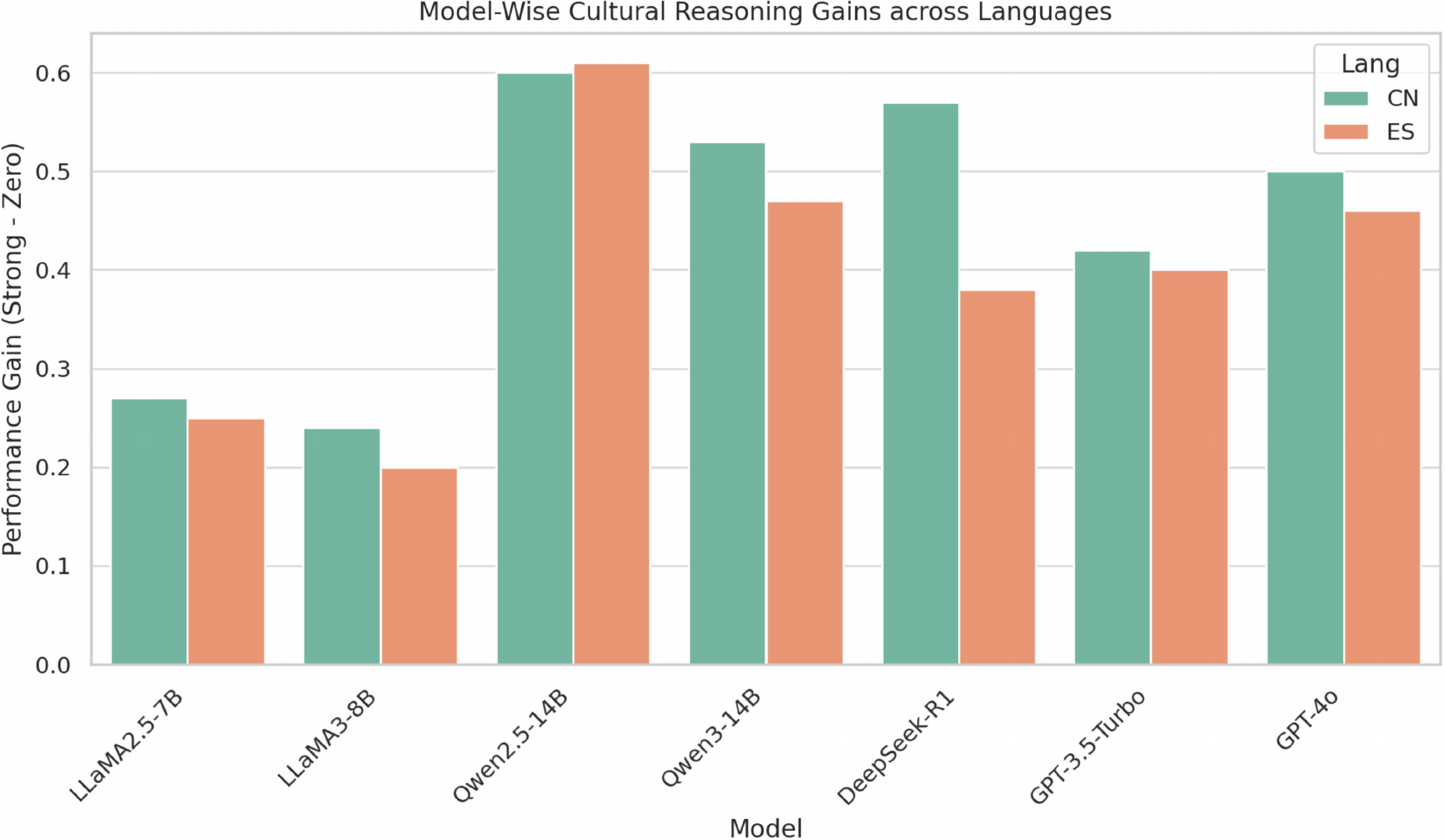}
\caption{Model-wise Reasoning Patterns}
\label{fig 4}
\end{figure*}

To further quantify each model’s adaptability to conceptual scaffolding, we calculate the performance gain from Zero to Strong injection levels across both Chinese and Spanish. As illustrated in Fig.~\ref{fig 4}, injection-induced gains vary considerably by model family and language, reflecting heterogeneous cultural interpretive capabilities.

\textbf{Qwen-series models} exhibit the steepest gains across the board. Qwen2.5-14B achieves the largest increase—over \textbf{0.60} points in both CN and ES—making it the most sensitive model to concept-level prompting. This highlights its strong ability to align with cultural frameworks when given sufficient guidance, despite weaker zero-shot performance.

\textbf{DeepSeek-R1} also performs impressively, with a gain of \textbf{0.57} in CN and \textbf{0.38} in ES. The cross-lingual variation suggests it generalizes more easily within its native (Chinese) linguistic-cultural context but encounters challenges with distant cultures like Spanish.

\textbf{GPT-4o} maintains the most balanced profile: while its gain is slightly lower (around \textbf{0.46–0.50}), its absolute performance remains top-tier. This indicates it requires less injection to perform well and already encodes substantial cultural generalization in its pretraining.

\textbf{GPT-3.5-Turbo} shows moderate sensitivity (\textbf{0.40–0.42 gain}), confirming it benefits from injection but still lags behind GPT-4o in interpretive versatility.

By contrast, \textbf{LLaMA-based models} are consistently the least responsive. LLaMA2.5-7B and LLaMA3-8B show modest gain ranges (\textbf{0.20–0.27}) and remain under 0.35 in absolute performance. This plateau suggests their training regimes and alignment tuning lack sufficient cultural depth or symbolic interpretive capacity, limiting their growth under concept-driven scenarios.

\subsection{E2: Implication}

These results reinforce the idea that \textbf{conceptual injection acts as a diagnostic lens}: models that exhibit large gain curves benefit from structured guidance, while those with flatter gains either already encode relevant knowledge (like GPT-4o) or lack the capacity to utilize it (like LLaMA). This performance gain analysis not only helps rank LLMs by cultural reasoning adaptability but also suggests future directions in alignment and training strategies for cross-cultural robustness.

\section{Appendix F: Dimension-wise Analysis}

\subsection{Metaphor and Symbolism}

In Chinese, GPT-4o already performs strongly in zero-shot settings (0.45), and reaches near-perfect accuracy under strong prompts (0.93), suggesting that metaphorical understanding benefits from both prior exposure and contextual reinforcement. In Spanish, however, the zero-shot baseline drops to 0.38, underscoring challenges in transferring metaphor schemas across distant cultural spaces. Strong prompting improves this to 0.86, confirming injection's effectiveness in activating symbolic interpretive capacities.

\begin{figure*}[t]
    \centering
    \includegraphics[width=0.95\linewidth]{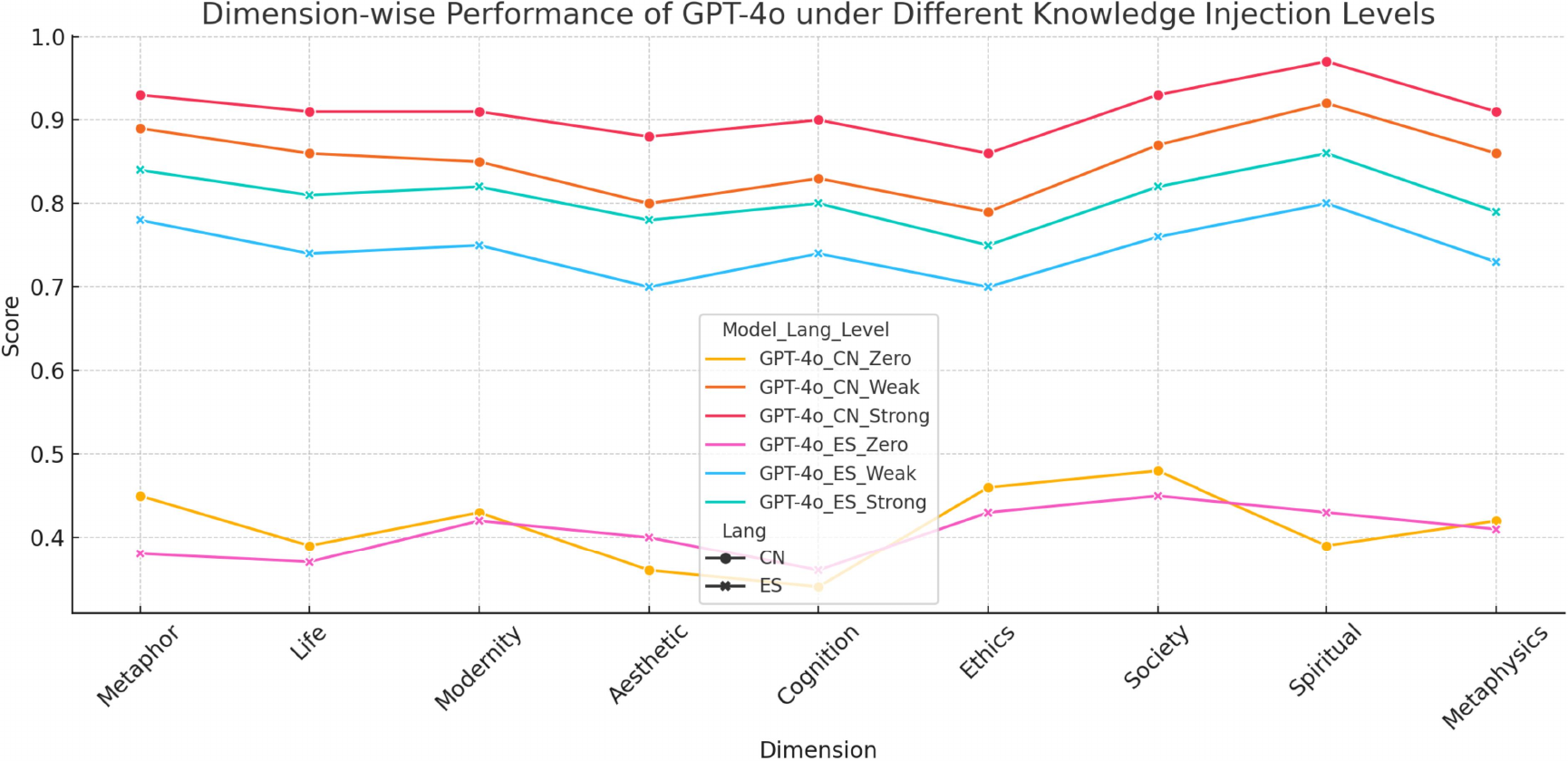}
    \caption{Dimension-wise performance of GPT-4o under different levels of conceptual knowledge injection in Chinese (CN) and Spanish (ES). Results are shown across nine cultural dimensions, revealing injection responsiveness and cross-lingual generalization patterns.}
    \label{fig 5}
\end{figure*}

Fig.~\ref{fig 5} illustrates GPT-4o’s performance on each of the nine cultural dimensions under varying levels of conceptual injection, in both Chinese (CN) and Spanish (ES). The results reveal notable differences in dimension-level sensitivity, baseline capability, and injection responsiveness.

\subsection{Life and Practices}

This dimension shows the lowest baseline performance across both languages. In zero-shot conditions, scores hover below 0.40 (CN: 0.39; ES: 0.37), indicating that culturally situated daily behaviors—such as festivals, food customs, or etiquette—are hard to infer without explicit cues. Even under strong injection, performance remains relatively lower compared to other domains, suggesting a deeper need for culturally embedded training data.

\subsection{Modernity and Individuality}

Moderate baseline scores (CN: 0.43; ES: 0.42) suggest some generalizable understanding of modernity-related concepts such as progress, rights, or individuality. However, a clear injection gradient appears—rising steadily toward 0.91 (CN) and 0.86 (ES)—implying that abstract values require reinforcement through contextual signals to be interpreted in culturally congruent ways.

\subsection{Aesthetic and Artistic Values}

Performance in this domain remains the most fragile. Both CN and ES zero-shot scores fall below 0.40, and even strong prompts yield lower final scores than other dimensions (CN: 0.88; ES: 0.79). The limited gain reflects the subjective, culturally coded nature of aesthetic preferences and symbolic expressions, which are less likely to be aligned without deep cultural modeling.

\subsection{Cognition and Thinking Patterns}

GPT-4o shows relatively modest performance here, with Chinese zero-shot at 0.36 and Spanish at 0.35. Injection prompts yield steady improvements (CN: 0.90; ES: 0.80), but gains are smaller than in ethical or social domains. This suggests that cognitive metaphors and epistemological habits (e.g., linear vs. holistic thinking) remain challenging to model without domain-specific examples.

\subsection{Ethics and Values}

This dimension reveals strong gains from injection. Chinese performance moves from 0.46 (Zero) to 0.94 (Strong), while Spanish rises from 0.44 to 0.86. The steep curve implies that ethical reasoning is sensitive to cultural conceptual framing and benefits greatly from prompt-level grounding, making it a useful dimension for probing moral universals and divergences.

\subsection{Society and Social Order}

Zero-injection scores are modest but consistent (CN: 0.48; ES: 0.46). After strong injection, both reach high levels (CN: 0.92; ES: 0.85). This indicates that LLMs can generalize societal structures (e.g., authority, citizenship, power) across cultures when provided with aligned concepts, and perform well in comparative sociocultural contexts.

\subsection{Spirituality and Religion}

This is one of GPT-4o’s strongest dimensions in Chinese (0.39 → 0.97). Spanish performance also improves markedly (0.42 → 0.86), showing GPT-4o's ability to navigate symbolic and ritualistic content with prompt guidance. The steep growth reflects the potential of injection to scaffold complex metaphysical reasoning grounded in ritual or belief systems.

\subsection{Metaphysics and Cosmology}

Both CN and ES start from moderate levels (~0.42) and rise to high performance under strong injection (CN: 0.91; ES: 0.79). The dimensional trend suggests that GPT-4o can handle abstract ontological concepts effectively once given culturally specific anchors (e.g., yin-yang, reincarnation, etc.).

\subsection{Summary Insights}

Across all nine dimensions, GPT-4o exhibits consistent benefit from conceptual injection, with the steepest improvements observed in \textit{Spirituality}, \textit{Ethics}, and \textit{Metaphysics}—domains that require symbolic, abstract, and culturally rooted reasoning. Dimensions like \textit{Aesthetic Values} and \textit{Life Practices}, by contrast, remain more resistant to prompt-based enhancement, likely due to their subjective and context-heavy nature. Overall, this dimensional breakdown highlights the necessity of fine-grained evaluation when assessing LLMs' cross-cultural capacities.

\section{Appendix G: Injection Sensitivity}

\begin{figure*}[t]
\centering
\includegraphics[width=0.95\textwidth]{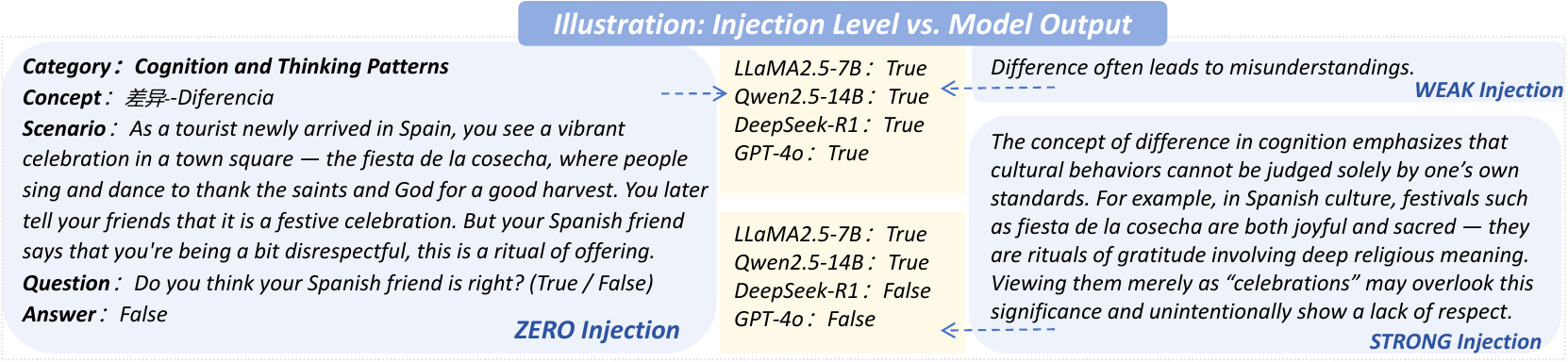}
\caption{Case analysis. It is important to note that the data in the dataset was originally written in Spanish; here, it has been translated into English for ease of reading. This is a typical error case: under the zero-injection setting, all LLMs produced incorrect answers, as the scenario is highly prone to cultural ambiguity and stereotypical reasoning. Even under the weak-injection setting, the models still failed. Only when a detailed explanation of the cross-cultural core concept in this specific context was provided (strong injection) did the models give correct responses.}
\label{fig 6}
\end{figure*}

Prompt examples for the three concept injection levels are shown in Fig.~\ref{fig 6}, and the corresponding results are reported in Table~3.

\begin{table*}[t]
  \centering
  \scriptsize
  \resizebox{1\textwidth}{!}{%
\begin{tabular}{cccccccccccc}
\hline
\textbf{Model} & \textbf{Lang} & \textbf{Metaphor} & \textbf{Life} & \textbf{Modernity} & \textbf{Aesthetic} & \textbf{Cognition} & \textbf{Ethics} & \textbf{Society} & \textbf{Spiritual} & \textbf{Metaphysics} & \textbf{Average} \\
\hline

\multicolumn{12}{c}{\textit{\textbf{Zero Conceptual Knowledge Injection}}} \\
\hline
{LLaMA2.5-7B} & CN/ES & 0.06/0.03 & 0.04/0.01 & 0.06/0.04 & 0.02/0.02 & 0.02/0.01 & 0.05/0.04 & 0.08/0.05 & 0.03/0.05 & 0.02/0.05 & 0.04/0.03 \\
{LLaMA3-8B}  & CN/ES   & 0.13/0.05 & 0.13/0.04 & 0.21/0.25 & 0.08/0.23 & 0.11/0.25 & 0.05/0.13 & 0.04/0.09 & 0.04/0.11 & 0.14/0.05 & 0.10/0.13 \\
{Qwen2.5-14B} & CN/ES  & 0.15/0.07 & 0.11/0.05 & 0.22/0.26 & 0.09/0.24 & 0.12/0.26 & 0.06/0.14 & 0.05/0.11 & 0.05/0.12 & 0.15/0.06 & 0.11/0.15 \\
{Qwen3-14B}  & CN/ES   & 0.26/0.35 & 0.14/0.37 & 0.23/0.35 & 0.26/0.33 & 0.17/0.23 & 0.23/0.35 & 0.33/0.43 & 0.33/0.36 & 0.18/0.14 & 0.23/0.32 \\
{DeepSeek-R1} & CN/ES  & 0.40/0.22 & 0.38/0.24 & 0.42/0.25 & 0.41/0.26 & 0.39/0.23 & 0.35/0.21 & 0.41/0.28 & 0.43/0.27 & 0.39/0.24 & 0.40/0.25 \\
{GPT-3.5-Turbo} & CN/ES & 0.41/0.36 & 0.36/0.40 & 0.40/0.41 & 0.34/0.38 & 0.35/0.37 & 0.43/0.41 & 0.39/0.44 & 0.37/0.40 & 0.38/0.39 & 0.38/0.39 \\
{GPT-4o}  & CN/ES      & 0.45/0.38 & 0.39/0.37 & 0.43/0.42 & 0.36/0.40 & 0.34/0.36 & 0.46/0.43 & 0.48/0.45 & 0.39/0.43 & 0.42/0.41 & 0.41/0.40 \\

\hline
\multicolumn{12}{c}{\textit{\textbf{Weak Conceptual Knowledge Injection}}} \\
\hline
{LLaMA2.5-7B} & CN & 0.23 {\tiny (×0.26)} & 0.21 {\tiny (×0.19)} & 0.25 {\tiny (×0.24)} & 0.22 {\tiny (×0.09)} & 0.23 {\tiny (×0.09)} & 0.20 {\tiny (×0.25)} & 0.24 {\tiny (×0.33)} & 0.25 {\tiny (×0.12)} & 0.22 {\tiny (×0.09)} & 0.23 {\tiny (×0.17)} \\
              & ES & 0.20 {\tiny (×0.15)} & 0.18 {\tiny (×0.06)} & 0.22 {\tiny (×0.18)} & 0.19 {\tiny (×0.11)} & 0.20 {\tiny (×0.05)} & 0.18 {\tiny (×0.22)} & 0.20 {\tiny (×0.25)} & 0.22 {\tiny (×0.23)} & 0.19 {\tiny (×0.26)} & 0.20 {\tiny (×0.15)} \\

{LLaMA3-8B} & CN & 0.36 {\tiny (×0.36)} & 0.35 {\tiny (×0.37)} & 0.38 {\tiny (×0.55)} & 0.40 {\tiny (×0.20)} & 0.39 {\tiny (×0.28)} & 0.33 {\tiny (×0.15)} & 0.39 {\tiny (×0.10)} & 0.42 {\tiny (×0.10)} & 0.38 {\tiny (×0.37)} & 0.38 {\tiny (×0.26)} \\
            & ES & 0.31 {\tiny (×0.16)} & 0.30 {\tiny (×0.13)} & 0.36 {\tiny (×0.69)} & 0.28 {\tiny (×0.82)} & 0.35 {\tiny (×0.71)} & 0.28 {\tiny (×0.46)} & 0.32 {\tiny (×0.28)} & 0.40 {\tiny (×0.28)} & 0.28 {\tiny (×0.18)} & 0.32 {\tiny (×0.41)} \\

{Qwen2.5-14B} & CN & 0.48 {\tiny (×0.31)} & 0.52 {\tiny (×0.21)} & 0.48 {\tiny (×0.46)} & 0.42 {\tiny (×0.21)} & 0.50 {\tiny (×0.24)} & 0.45 {\tiny (×0.13)} & 0.49 {\tiny (×0.10)} & 0.53 {\tiny (×0.09)} & 0.48 {\tiny (×0.31)} & 0.48 {\tiny (×0.23)} \\
              & ES & 0.45 {\tiny (×0.16)} & 0.38 {\tiny (×0.13)} & 0.43 {\tiny (×0.60)} & 0.35 {\tiny (×0.69)} & 0.38 {\tiny (×0.68)} & 0.33 {\tiny (×0.42)} & 0.37 {\tiny (×0.30)} & 0.55 {\tiny (×0.22)} & 0.36 {\tiny (×0.17)} & 0.40 {\tiny (×0.38)} \\

{Qwen3-14B} & CN & 0.54 {\tiny (×0.48)} & 0.60 {\tiny (×0.23)} & 0.55 {\tiny (×0.42)} & 0.46 {\tiny (×0.57)} & 0.59 {\tiny (×0.29)} & 0.53 {\tiny (×0.43)} & 0.57 {\tiny (×0.58)} & 0.62 {\tiny (×0.53)} & 0.56 {\tiny (×0.32)} & 0.55 {\tiny (×0.42)} \\
             & ES & 0.50 {\tiny (×0.70)} & 0.43 {\tiny (×0.86)} & 0.49 {\tiny (×0.71)} & 0.42 {\tiny (×0.79)} & 0.45 {\tiny (×0.51)} & 0.40 {\tiny (×0.87)} & 0.43 {\tiny (×0.79)} & 0.65 {\tiny (×0.55)} & 0.43 {\tiny (×0.33)} & 0.48 {\tiny (×0.67)} \\

{DeepSeek-R1} & CN & 0.52 {\tiny (×0.77)} & 0.50 {\tiny (×0.76)} & 0.49 {\tiny (×0.86)} & 0.46 {\tiny (×0.89)} & 0.48 {\tiny (×0.81)} & 0.44 {\tiny (×0.80)} & 0.50 {\tiny (×0.82)} & 0.54 {\tiny (×0.80)} & 0.50 {\tiny (×0.78)} & 0.49 {\tiny (×0.82)} \\
              & ES & 0.38 {\tiny (×0.58)} & 0.36 {\tiny (×0.67)} & 0.37 {\tiny (×0.68)} & 0.34 {\tiny (×0.76)} & 0.36 {\tiny (×0.64)} & 0.32 {\tiny (×0.66)} & 0.35 {\tiny (×0.80)} & 0.37 {\tiny (×0.73)} & 0.34 {\tiny (×0.74)} & 0.35 {\tiny (×0.71)} \\

{GPT-3.5-Turbo} & CN & 0.85 {\tiny (×0.48)} & 0.83 {\tiny (×0.43)} & 0.81 {\tiny (×0.49)} & 0.77 {\tiny (×0.44)} & 0.79 {\tiny (×0.44)} & 0.75 {\tiny (×0.57)} & 0.84 {\tiny (×0.46)} & 0.89 {\tiny (×0.41)} & 0.83 {\tiny (×0.46)} & 0.82 {\tiny (×0.46)} \\
                & ES & 0.75 {\tiny (×0.48)} & 0.70 {\tiny (×0.57)} & 0.72 {\tiny (×0.57)} & 0.68 {\tiny (×0.56)} & 0.71 {\tiny (×0.52)} & 0.65 {\tiny (×0.63)} & 0.73 {\tiny (×0.57)} & 0.77 {\tiny (×0.52)} & 0.70 {\tiny (×0.56)} & 0.72 {\tiny (×0.53)} \\

{GPT-4o} & CN & 0.89 {\tiny (×0.51)} & 0.86 {\tiny (×0.45)} & 0.85 {\tiny (×0.51)} & 0.80 {\tiny (×0.45)} & 0.83 {\tiny (×0.41)} & 0.79 {\tiny (×0.58)} & 0.87 {\tiny (×0.55)} & 0.92 {\tiny (×0.42)} & 0.86 {\tiny (×0.49)} & 0.85 {\tiny (×0.48)} \\
         & ES & 0.78 {\tiny (×0.49)} & 0.74 {\tiny (×0.50)} & 0.75 {\tiny (×0.56)} & 0.70 {\tiny (×0.57)} & 0.74 {\tiny (×0.49)} & 0.70 {\tiny (×0.61)} & 0.76 {\tiny (×0.50)} & 0.80 {\tiny (×0.63)} & 0.73 {\tiny (×0.41)} & 0.75 {\tiny (×0.53)} \\

\hline
\multicolumn{12}{c}{\textit{\textbf{Strong Conceptual Knowledge Injection}}} \\
\hline

{LLaMA2.5-7B} & CN & 0.32 {\tiny (×0.19)} & 0.30 {\tiny (×0.13)} & 0.34 {\tiny (×0.18)} & 0.29 {\tiny (×0.07)} & 0.33 {\tiny (×0.06)} & 0.28 {\tiny (×0.18)} & 0.34 {\tiny (×0.24)} & 0.36 {\tiny (×0.08)} & 0.31 {\tiny (×0.06)} & 0.31 {\tiny (×0.13)} \\
              & ES & 0.29 {\tiny (×0.10)} & 0.27 {\tiny (×0.04)} & 0.31 {\tiny (×0.13)} & 0.26 {\tiny (×0.08)} & 0.28 {\tiny (×0.04)} & 0.25 {\tiny (×0.16)} & 0.29 {\tiny (×0.17)} & 0.32 {\tiny (×0.16)} & 0.28 {\tiny (×0.18)} & 0.28 {\tiny (×0.11)} \\

{LLaMA3-8B} & CN & 0.35 {\tiny (×0.37)} & 0.33 {\tiny (×0.39)} & 0.38 {\tiny (×0.55)} & 0.31 {\tiny (×0.26)} & 0.37 {\tiny (×0.30)} & 0.29 {\tiny (×0.17)} & 0.39 {\tiny (×0.10)} & 0.40 {\tiny (×0.10)} & 0.35 {\tiny (×0.40)} & 0.35 {\tiny (×0.29)} \\
            & ES & 0.32 {\tiny (×0.16)} & 0.30 {\tiny (×0.13)} & 0.35 {\tiny (×0.71)} & 0.28 {\tiny (×0.82)} & 0.33 {\tiny (×0.76)} & 0.26 {\tiny (×0.50)} & 0.32 {\tiny (×0.28)} & 0.38 {\tiny (×0.29)} & 0.31 {\tiny (×0.16)} & 0.31 {\tiny (×0.42)} \\

{Qwen2.5-14B} & CN & 0.74 {\tiny (×0.20)} & 0.81 {\tiny (×0.14)} & 0.74 {\tiny (×0.30)} & 0.61 {\tiny (×0.15)} & 0.80 {\tiny (×0.15)} & 0.70 {\tiny (×0.09)} & 0.76 {\tiny (×0.07)} & 0.85 {\tiny (×0.06)} & 0.78 {\tiny (×0.19)} & 0.75 {\tiny (×0.15)} \\
              & ES & 0.72 {\tiny (×0.10)} & 0.60 {\tiny (×0.08)} & 0.68 {\tiny (×0.38)} & 0.49 {\tiny (×0.49)} & 0.56 {\tiny (×0.46)} & 0.46 {\tiny (×0.30)} & 0.54 {\tiny (×0.20)} & 0.87 {\tiny (×0.14)} & 0.58 {\tiny (×0.10)} & 0.61 {\tiny (×0.25)} \\

{Qwen3-14B} & CN & 0.88 {\tiny (×0.30)} & 0.94 {\tiny (×0.15)} & 0.87 {\tiny (×0.26)} & 0.77 {\tiny (×0.34)} & 0.92 {\tiny (×0.19)} & 0.85 {\tiny (×0.27)} & 0.91 {\tiny (×0.36)} & 0.98 {\tiny (×0.34)} & 0.90 {\tiny (×0.20)} & 0.89 {\tiny (×0.26)} \\
             & ES & 0.83 {\tiny (×0.42)} & 0.75 {\tiny (×0.49)} & 0.81 {\tiny (×0.43)} & 0.67 {\tiny (×0.49)} & 0.73 {\tiny (×0.32)} & 0.63 {\tiny (×0.56)} & 0.72 {\tiny (×0.60)} & 0.99 {\tiny (×0.36)} & 0.70 {\tiny (×0.20)} & 0.76 {\tiny (×0.42)} \\

{DeepSeek-R1} & CN & 0.87 {\tiny (×0.46)} & 0.84 {\tiny (×0.45)} & 0.83 {\tiny (×0.51)} & 0.81 {\tiny (×0.51)} & 0.85 {\tiny (×0.46)} & 0.79 {\tiny (×0.44)} & 0.88 {\tiny (×0.47)} & 0.93 {\tiny (×0.46)} & 0.87 {\tiny (×0.45)} & 0.85 {\tiny (×0.47)} \\
              & ES & 0.68 {\tiny (×0.32)} & 0.62 {\tiny (×0.39)} & 0.65 {\tiny (×0.38)} & 0.60 {\tiny (×0.43)} & 0.63 {\tiny (×0.37)} & 0.56 {\tiny (×0.38)} & 0.63 {\tiny (×0.44)} & 0.66 {\tiny (×0.41)} & 0.61 {\tiny (×0.39)} & 0.62 {\tiny (×0.40)} \\

{GPT-3.5-Turbo} & CN & 0.90 {\tiny (×0.46)} & 0.89 {\tiny (×0.40)} & 0.88 {\tiny (×0.46)} & 0.85 {\tiny (×0.40)} & 0.87 {\tiny (×0.40)} & 0.82 {\tiny (×0.52)} & 0.91 {\tiny (×0.43)} & 0.95 {\tiny (×0.39)} & 0.89 {\tiny (×0.43)} & 0.89 {\tiny (×0.43)} \\
                & ES & 0.81 {\tiny (×0.44)} & 0.77 {\tiny (×0.52)} & 0.78 {\tiny (×0.53)} & 0.74 {\tiny (×0.51)} & 0.76 {\tiny (×0.49)} & 0.71 {\tiny (×0.58)} & 0.79 {\tiny (×0.56)} & 0.82 {\tiny (×0.49)} & 0.76 {\tiny (×0.51)} & 0.78 {\tiny (×0.50)} \\

{GPT-4o} & CN & 0.93 {\tiny (×0.48)} & 0.91 {\tiny (×0.43)} & 0.91 {\tiny (×0.47)} & 0.88 {\tiny (×0.41)} & 0.90 {\tiny (×0.38)} & 0.86 {\tiny (×0.53)} & 0.93 {\tiny (×0.52)} & 0.97 {\tiny (×0.40)} & 0.91 {\tiny (×0.46)} & 0.91 {\tiny (×0.45)} \\
         & ES & 0.84 {\tiny (×0.45)} & 0.81 {\tiny (×0.46)} & 0.82 {\tiny (×0.51)} & 0.78 {\tiny (×0.51)} & 0.80 {\tiny (×0.45)} & 0.75 {\tiny (×0.57)} & 0.82 {\tiny (×0.55)} & 0.86 {\tiny (×0.50)} & 0.79 {\tiny (×0.52)} & 0.81 {\tiny (×0.49)} \\

\hline
\end{tabular}%
}
\caption{Performance of LLMs on cross-cultural understanding and reasoning in real-world scenarios is summarized, with cross-cultural capabilities categorized into nine dimensions. The impact of concept knowledge injection is evaluated across three levels: zero, weak, and strong. Numbers in parentheses represent multiplicative improvements relative to the zero injection baseline.}
\label{table 3}
\end{table*}

\subsection{Overall Injection Effects across Languages}

\begin{figure*}[ht]
\centering
\includegraphics[width=0.95\textwidth]{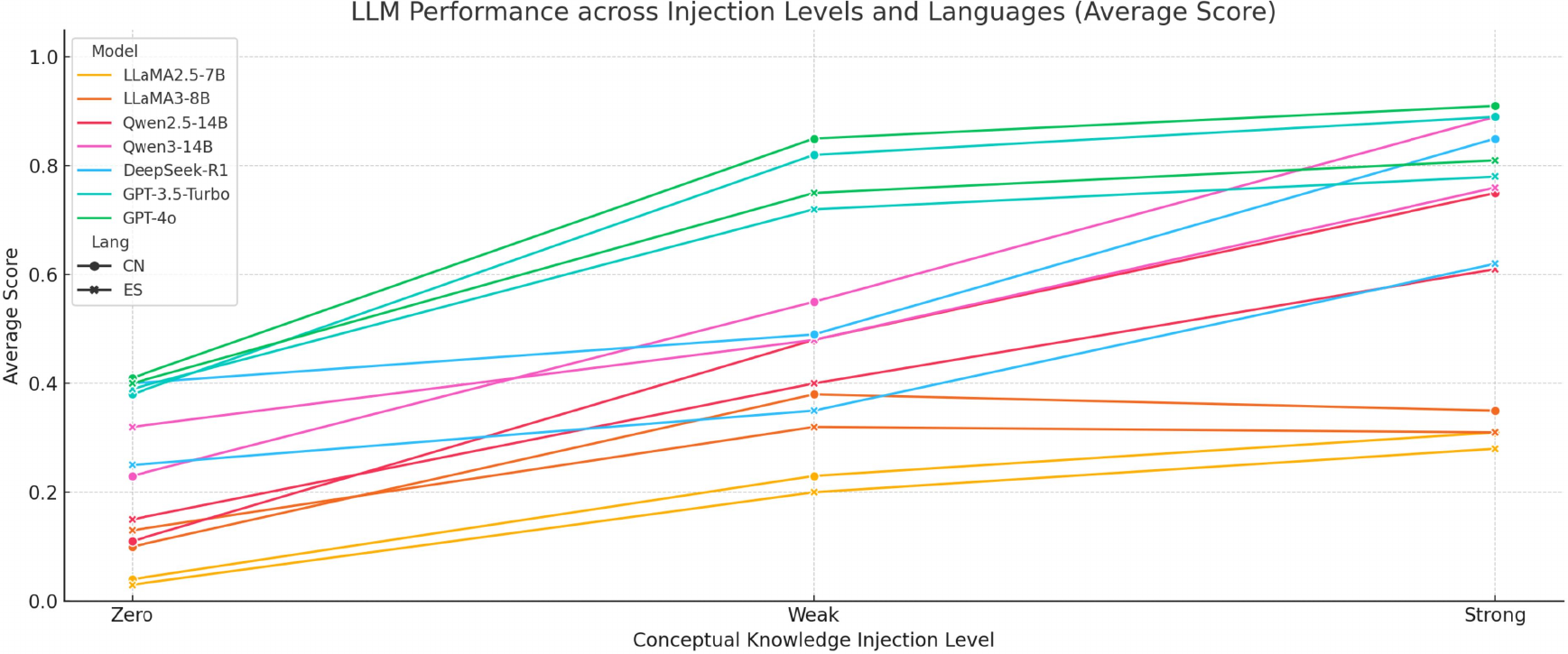}
\caption{Average performance of seven LLMs across conceptual injection levels in Chinese and Spanish.}
\label{fig 7}
\end{figure*}

To assess the impact of conceptual knowledge injection on cross-cultural reasoning, we evaluated seven representative LLMs across three injection levels (Zero, Weak, Strong) in both Chinese (CN) and Spanish (ES). Fig.~\ref{fig 7} reveals a consistent upward trajectory for most models, confirming that knowledge scaffolding substantially enhances model performance across languages.

GPT-4o demonstrates the highest performance across all injection levels, achieving an average score of 0.91 (CN) and 0.86 (ES) under strong prompting. GPT-3.5-Turbo and DeepSeek-R1 also exhibit stable gains. Meanwhile, Qwen2.5-14B and Qwen3-14B show sharper improvement from weak to strong levels—especially in Spanish—suggesting higher sensitivity to cultural grounding.

In contrast, LLaMA-based models plateau at lower performance ranges. LLaMA2.5-7B, in particular, shows only marginal gains between weak and strong levels, with its ES performance remaining under 0.30 even after full injection. This highlights the limited cultural interpretive capacity of smaller, instruction-tuned models under minimal guidance.

Overall, injection effects are more pronounced in ES than CN. This suggests that cultural distance (e.g., Chinese-to-Spanish transfer) increases reliance on explicit conceptual grounding. These trends validate the efficacy of the SAGE benchmark’s injection framework in scaling interpretive difficulty and probing cultural reasoning depth.

\subsection{High-Capacity Models: Robustness and Generalization}

Among all evaluated systems, GPT-4o shows both high baseline capability and strong injection adaptability. Despite already achieving near 0.40 accuracy under Zero injection, it gains over 0.50 in both CN and ES with strong prompts, reaching top performance. This underscores GPT-4o's superior cultural generalization, even in low-context settings.

GPT-3.5-Turbo follows a more modest growth pattern, maintaining decent performance throughout. Although its gain is less dramatic than GPT-4o’s, its consistency across languages and injection levels reflects a well-balanced architecture with moderate cultural adaptability.

DeepSeek-R1 occupies a strong middle ground. It starts from a higher zero-shot baseline than Qwen models and improves steadily across injection levels. This smooth gain profile indicates lower dependence on external concept scaffolding and possibly stronger internal alignment with culturally grounded semantics.

\subsection{Mid-Tier Models: Injection-Sensitive Improvements}

Qwen2.5-14B and Qwen3-14B present steep growth curves, especially in Spanish. Their performance at Zero injection is notably low (around 0.15–0.35), but rises sharply to 0.75–0.78 with full concept injection. This pronounced jump implies a high sensitivity to cultural framing and a dependence on explicit guidance for accurate reasoning.

This also suggests that Qwen models may lack pre-trained grounding in global cultural concepts but possess the architectural flexibility to respond well when such guidance is injected. Their strong responsiveness indicates potential for domain adaptation, though it also highlights limitations under low-resource settings.

\subsection{Low-End Models: Limited Cultural Adaptability}

LLaMA2.5-7B and LLaMA3-8B consistently occupy the lowest performance tier, both in Chinese and Spanish. Even under strong injection, their scores remain below 0.30. More importantly, the improvement from Zero to Strong is minimal, suggesting limited internal representation of metaphorical, symbolic, and culturally embedded reasoning.

These results reveal that model scale alone does not ensure cultural competence. Instruction tuning quality and data diversity play a more critical role, especially in tasks requiring nuanced cross-cultural interpretation.

\subsection{Takeaway: Injection Amplifies Cultural Distance Sensitivity}

Taken together, the model-wise differences confirm that \textit{conceptual injection acts as a cultural enabler}, especially in culturally distant settings like Chinese–Spanish transfer. The sharper injection-driven improvements in ES across models further imply that models struggle more without prior exposure to the target culture’s symbolic structures.

Therefore, injection sensitivity not only reflects model capacity but also indirectly quantifies its baseline cultural bias and adaptability. These findings support the role of SAGE injection as a scalable probe for cultural interpretability.

\section{Appendix H: Cross-Lingual Extension Results}

To validate the cross-lingual extensibility of the SAGE benchmark, we extend our evaluation to Korean (KO), leveraging two advanced open-source models: Qwen2.5-14B and DeepSeek-R1. Both models were tested under consistent concept-injection settings, and performance was measured across six cultural dimensions:

\begin{itemize}
    \item \textbf{MS}: Metaphor and Symbol
    \item \textbf{LP}: Life Practices
    \item \textbf{AAV}: Aesthetic and Artistic Values
    \item \textbf{EV}: Ethical Values
    \item \textbf{SSO}: Society and Social Order
    \item \textbf{SR}: Spirituality and Rituals
\end{itemize}

\begin{table}[t]
\centering
\scriptsize
\begin{tabular}{cccccccc}
\hline
\textbf{Model} & \textbf{Lang} & \textbf{MS} & \textbf{LP} & \textbf{AAV} & \textbf{EV} & \textbf{SSO} & \textbf{SR} \\
\hline
Qwen2.5-14B & CN & 0.86 & 0.94 & 0.66 & 0.88 & 0.86 & 0.97 \\
            & KO & 0.84 & 0.87 & 0.56 & 0.66 & 0.64 & 0.94 \\
\hline
DeepSeek-R1 & CN & 0.92 & 0.97 & 0.86 & 0.86 & 0.91 & 0.97 \\
            & KO & 0.88 & 0.91 & 0.56 & 0.72 & 0.82 & 0.91 \\
\hline
\end{tabular}
\caption{Cross-lingual performance comparison across six cultural dimensions (Chinese vs. Korean)}
\label{table 4}
\end{table}

As shown in Table~4, both models maintained relatively strong performance when extended to Korean. Consistency was particularly notable in \textbf{Spirituality and Rituals (SR)}, \textbf{Life Practices (LP)}, and \textbf{Metaphor and Symbol (MS)}, suggesting that dimensions grounded in shared East Asian philosophical and ritual traditions are more amenable to cross-lingual transfer.

In contrast, \textbf{Aesthetic and Artistic Values (AAV)} exhibited a sharp decline in Korean, especially for Qwen2.5-14B, with a drop of 10 percentage points compared to its Chinese performance. This suggests that cultural expressions relying heavily on local aesthetic codes are more challenging to transfer—even between script-divergent but culturally proximate languages.

\begin{figure}[t]
\centering
\includegraphics[width=0.9\columnwidth]{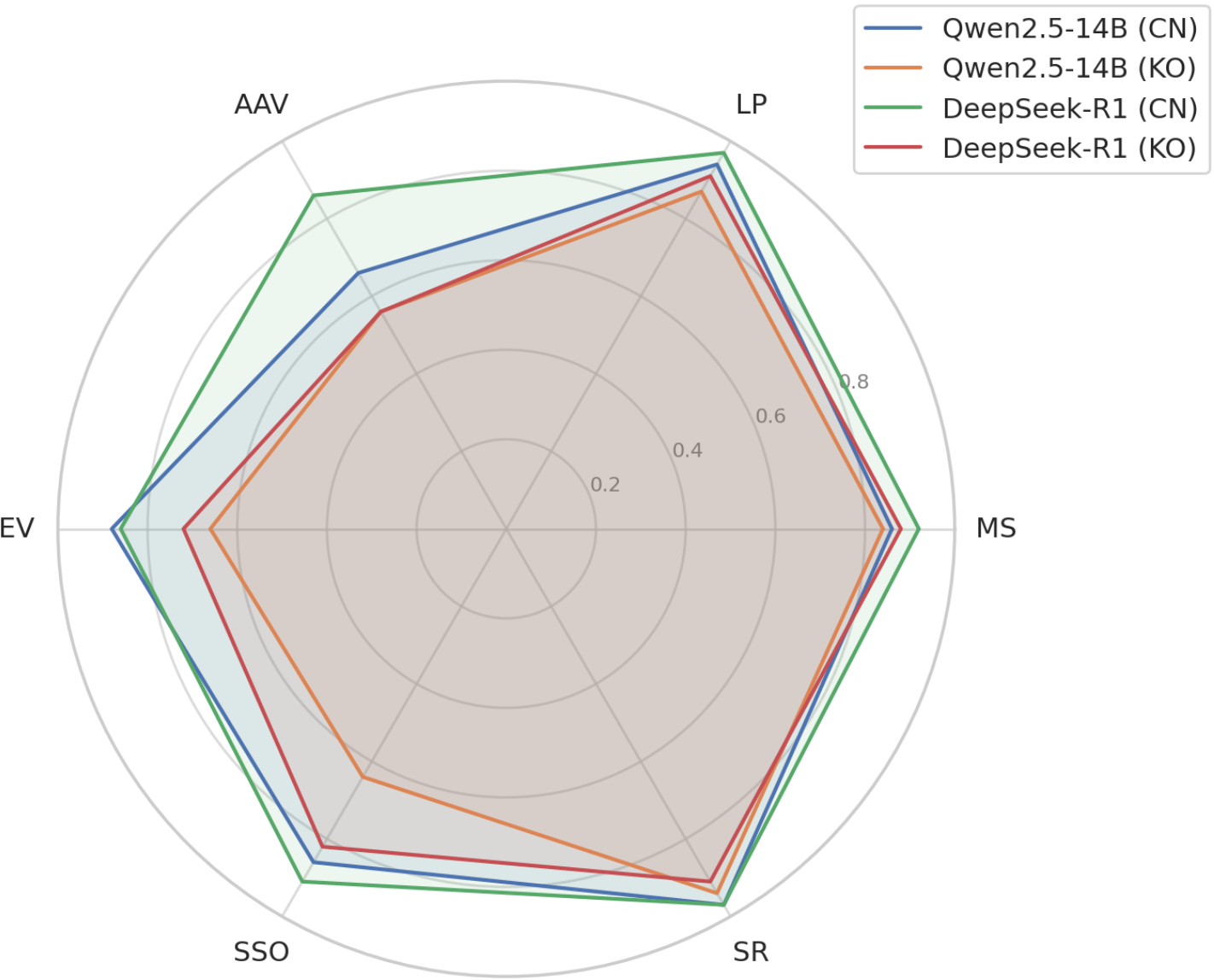}
\caption{Radar plot revealing dimension-level transfer stability from Chinese to Korean across six cultural domains}
\label{fig 8}
\end{figure}

Fig.~\ref{fig 8} further illustrates the above trends through a radar visualization. While Chinese consistently outperforms Korean on most dimensions, the gap is narrower in structural and ethical domains. Notably, DeepSeek-R1 maintains a more compact and stable radar shape across languages than Qwen2.5-14B, hinting at stronger alignment with culturally distributed representations. Overall, these results underscore the dimension-sensitive nature of cross-lingual cultural transfer, validating the flexibility and transferability of the SAGE benchmark.

To further assess robustness under different degrees of cultural context injection, Fig.~\ref{fig 8} presents performance across all nine cultural dimensions under \textbf{weak}, \textbf{medium}, and \textbf{strong} prompt conditions in Chinese (CN) and Spanish (ES). Overall, we observe a consistent upward trajectory as prompt richness increases, confirming the sensitivity of LLM performance to conceptual grounding.

Interestingly, \textbf{cross-lingual gaps diminish significantly under stronger prompts}. For instance, weak prompts in Spanish underperform across most dimensions, particularly in metaphor-heavy domains like \textbf{Metaphor and Symbolism} and \textbf{Spirituality and Religion}, highlighting the difficulty of symbolic transfer without contextual scaffolding. However, with strong prompts, Spanish performance closely approaches Chinese across nearly all domains—especially in \textbf{Modernity and Individuality}, \textbf{Cognition and Thinking Patterns}, and \textbf{Society and Social Order}.

This finding reinforces the idea that \textbf{cultural prompt engineering acts as an effective compensatory mechanism} for overcoming linguistic or cultural distance. It also validates the injection-controlled structure of SAGE as a means of exposing models' generalization boundaries and cultural flexibility.

\section{Criteria for Acceptable Questions}

To ensure the cultural and pragmatic validity of the dataset, we established a rigorous human review process based on well-defined acceptance criteria. A question was deemed \textit{acceptable} only if it fulfilled multiple dimensions of quality related to content, cultural fidelity, and reasoning design. These criteria were informed by the interdisciplinary nature of our benchmark, which integrates elements of intercultural communication, pragmatics, and metaphor theory.

\subsection{Cultural Authenticity and Contextual Relevance} 
Questions must reflect culturally grounded symbols, metaphors, or practices that are commonly recognizable and interpretable by native speakers of the target culture. Scenarios relying on vague, decontextualized, or pan-human symbols (e.g., “love,” “sun,” “happy”) without culture-specific framing were disqualified. Reviewers assessed whether the cultural cues were nuanced, avoiding stereotypes and tokenistic representations.

\subsection{Depth of Pragmatic or Emotional Reasoning}
Each item must go beyond mere cultural fact recall. The question should embed interpretive tension, emotional misalignment, symbolic ambiguity, or conflict in conversational context. This design enables evaluation of models' ability to infer intent, navigate implicit meanings, and resolve symbolic misunderstandings—core elements of cross-cultural sensitivity. Questions that only required surface-level identification (e.g., “What does a red envelope mean?”) were rejected.

\subsection{Inclusion of Culturally Meaningful Distractors (Trap Design)}
At least one distractor must represent a culturally plausible misreading or false inference. Trap types include:
\begin{itemize}
    \item \textit{Universalist traps:} assuming cultural symbols have the same meaning globally.
    \item \textit{Projection traps:} applying one’s own cultural emotional logic to another culture.
    \item \textit{Analogy mismatch:} incorrectly mapping foreign symbols to domestic equivalents.
    \item \textit{Tone mismatch:} culturally inappropriate conversational tone (e.g., too direct, confrontational).
\end{itemize}
Items with neutral or uninformative distractors (e.g., “None of the above”) were excluded.

\subsection{Conversational Plausibility and Naturalness}

Because questions are embedded in simulated dialogue or situated scenes, all utterances—both stems and options—must resemble natural language usage. Questions that sounded overly academic, mechanical, or generic were flagged for revision. Reviewers assessed tone, idiomaticity, and context appropriateness of language in each scenario.

\subsection{Clarity, Ambiguity Balance, and Interpretive Space}
Good questions strike a balance between being clearly structured and allowing multiple plausible readings that reveal differences in cultural logic. Scenarios that were too obscure, ambiguous without intent, or overly obvious were flagged. Our aim was to test cultural interpretive flexibility, not trick or confuse the model/test-taker arbitrarily.

\subsection{Non-Redundancy and Concept Coverage}

Each accepted question was checked against existing items to avoid near duplicates. We ensured that all nine cultural dimensions and over 200 symbolic concepts were adequately represented, minimizing redundancy while maximizing semantic coverage and difficulty diversity.

\subsection{Review Procedure and Thresholds}
 
Questions were first authored by culturally competent annotators, then passed through a three-round human review process. Each item was rated by two independent reviewers along three axes:
\begin{itemize}
    \item Cultural fidelity (Does it authentically represent cultural symbols?)
    \item Pragmatic depth (Does it test inference beyond knowledge?)
    \item Language quality (Is the expression natural and contextually appropriate?)
\end{itemize}
Each axis used a 5-point Likert scale (1 = very poor, 5 = excellent). Questions with an average score below 4.0 on any axis were either revised or excluded. Inter-annotator disagreements exceeding 1 point triggered a third-party adjudication. Only items with majority consensus and high inter-rater reliability (Fleiss’ $\kappa = 0.87$) were retained.

These rigorous standards ensured that each item in our benchmark is not only valid and culturally grounded, but also capable of distinguishing between surface-level cultural recall and deep reasoning under uncertainty—a crucial property for evaluating large language models’ true cross-cultural understanding.

\end{document}